\documentclass[conference]{IEEEtran}
\usepackage{cite}
\usepackage{amsmath,amssymb,amsfonts}
\usepackage{algorithmic}
\usepackage{graphicx}
\usepackage{textcomp}
\usepackage{xcolor}
\usepackage{comment}
\usepackage{amsmath,amssymb,mathtools} 
\usepackage{color}
\usepackage{graphicx}
\usepackage{subfig}
\usepackage{booktabs}
\usepackage{multirow}
\usepackage{url}
\usepackage[ruled,vlined]{algorithm2e}

\newcommand{\alg}{MINT}
\def\bbR{\mathbb R}
\def\BX{\mathbf{X}}\def\BY{\mathbf{Y}}
\def\BZ{\mathbf{Z}} 
\def\bbR{{\mathbb R}}
\def\bx{\mathbf x} \def\by{\mathbf y} \def\bz{\mathbf z}
\def\diy{\displaystyle} \def\bbE{{\mathbb E}} 
\def\rd{{\rm d}} 
\def\diy{\displaystyle}

\begin{document}
\title{MINT: Deep Network Compression via Mutual Information-based Neuron Trimming}

\author{\IEEEauthorblockN{Madan Ravi Ganesh}
\IEEEauthorblockA{EECS\\
University of Michigan\\
Ann Arbor, Michigan}
\and
\IEEEauthorblockN{Jason J. Corso}
\IEEEauthorblockA{EECS\\
University of Michigan\\
Ann Arbor, Michigan}
\and
\IEEEauthorblockN{Salimeh Yasaei Sekeh}
\IEEEauthorblockA{CSE\\
University of Maine\\
Orono, Maine}}


%


\maketitle

\begin{abstract}
Most approaches to deep neural network compression via pruning either evaluate a filter's importance using its weights or optimize an alternative objective function with sparsity constraints.
While these methods offer a useful way to approximate contributions from similar filters, they often either ignore the dependency between layers or solve a more difficult optimization objective than standard cross-entropy. 
Our method, Mutual Information-based Neuron Trimming (\alg), approaches deep compression via pruning by enforcing sparsity based on the strength of the relationship between filters of adjacent layers, across every pair of layers. 
The relationship is calculated using conditional geometric mutual information which evaluates the amount of similar information exchanged between the filters using a graph-based criterion.
When pruning a network, we ensure that retained filters contribute the majority of the information towards succeeding layers which ensures high performance.
Our novel approach outperforms existing state-of-the-art compression-via-pruning methods on the standard benchmarks for this task: MNIST, CIFAR-10, and ILSVRC2012, across a variety of network architectures.
In addition, we discuss our observations of a common denominator between our pruning methodology's response to adversarial attacks and calibration statistics when compared to the original network.
\end{abstract}


%
\IEEEpeerreviewmaketitle

\section{Introduction}
\label{sec:introduction}
Balancing the trade-off between the size of a deep network and achieving high performance is the most important constraint when designing deep neural networks (DNN) that can easily be translated to hardware.
Although deep learning yields remarkable performance in real-world problems like medical diagnosis~\cite{lee2017deep,abdel2016breast,anirudh2016lung}, autonomous vehicles~\cite{tinchev2019learning,marina2019deep,grigorescu2019neurotrajectory}, and others, they consume a large amount of memory and computational resources which limit their large-scale deployment.
With current state-of-the-art deep networks spanning hundreds of millions if not billions of parameters~\cite{real2019regularized,huang2019gpipe}, compressing them while maintaining high performance is challenging.


\begin{figure}[t]
    \centering
    \includegraphics[width=\columnwidth]{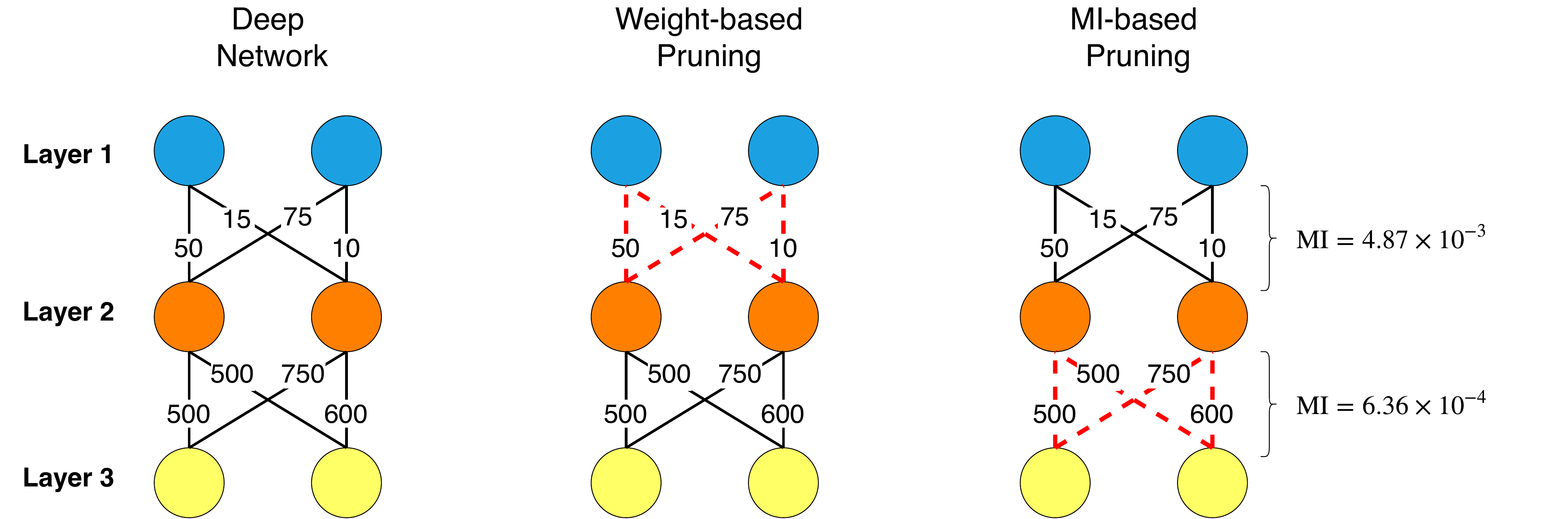}
    \caption{Weight-based pruning does not consider the dependency between layers. Instead it suggests the removal of low weight values. Mutual information (MI)-based pruning computes the value of information passed between layers, quantified by the MI value, and suggests the removal of higher weight values}
    \label{fig:mint_fig1}
\end{figure}

In this work, we approach DNN compression using network pruning~\cite{han2015learning}.
There are two broad approaches to network pruning, (a) unstructured pruning, where a filter's importance is evaluated using weights~\cite{han2015learning} or constraints like the $l_1$ norm~\cite{li2016pruning} on them, without considering the overall structure of sparsity induced, and (b) structured pruning, where the objective function is modified to include structured sparsity constraints~\cite{wen2016learning}.
Most structured pruning approaches ignore the dependency between layers and the impact of pruning on downstream layers while unstructured pruning methods force the network to optimize a harder and more sensitive optimization objective. 
The underlying common theme between both approaches is the use of filter weights as a proxy for their importance.


Evaluating a filter's importance purely from its weights is insufficient since it does not take into account the dependencies between filters or account for any form of uncertainty.
These factors are critical since higher weight values do not always represent its true importance and a filter's contribution can be compensated elsewhere in the network.
Consider the example shown in Fig.~\ref{fig:mint_fig1}, where a simple weight-based criterion suggests the removal of small valued weights.
However, the mutual information (MI) score, which we use to measure the dependency between pairs of filters and emphasize their importance, values the smaller weights over the large ones.
Pruning based on the MI scores would ensure a network where the retained filters pass on as much information as possible to the next layer.

\begin{figure*}[t]
    \centering
    \includegraphics[width=\textwidth]{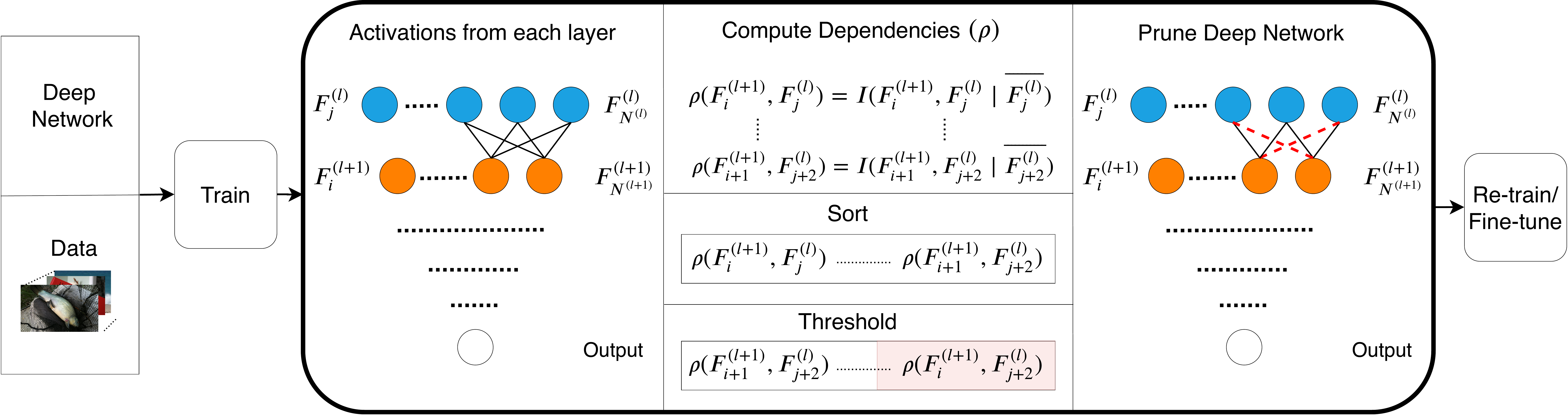}
    \caption{Illustration of the experimental setup highlighting the components of \alg. Between every pair of filters in consecutive layers ($l, l+1$) we compute the conditional geometric mutual information (GMI), using the activations from each filter, as the importance score. The total number of filters in each layer is defined by $N^{(l)}$ and $N^{(l+1)}$. The conditional GMI score indicates the importance of a filter in layer $l$'s contribution towards a filter in layer $l+1$. We then threshold filters based on the importance scores to ensure that we retain only filters that pass the majority of the information to successive layers. Finally, we retrain the network once to maintain a desired level of performance}
    \label{fig:mint_alg}
\end{figure*}

To overcome these issues, we propose Mutual Information-based Neuron Trimming (\alg) as a novel approach to compression-via-pruning in deep networks which stochastically accounts for the dependency between layers.
Fig.~\ref{fig:mint_alg} outlines our approach. 
In \alg, we use an estimator for conditional geometric mutual information (GMI), inspired by \cite{yasaei2019geometric}, to measure the dependency between filters of successive layers.
Specifically, we use a graph-based criterion (Friedman-Rafsky Statistic~\cite{friedman1983graph}) to measure the conditional GMI between the activations of filters at layer $l$ and $l+1$, denoted by $F^{(l)}_{i}, F^{(l+1)}_{j}$, given the remaining filters in layer $l$.
On evaluating all such dependencies, we sort the importance scores between filters of every pair of layers in the network.
Finally, we threshold a desired percentage of these values to retain filters that contribute the majority of the information to successive layers.
Hence, \alg~maintains high performance with compressed and retrained networks. 

Through \alg, we contribute a network pruning method that addresses the need to use dependency between layers as an important factor to prune networks. 
By maintaining filters that contribute the majority of the information passed between layers, we ensure that the impact of pruning on downstream layers is minimized.
In doing so, we achieve state-of-the-art performance across multiple Dataset-CNN architecture combinations, highlighting the general applicability of our method.

Further, we empirically analyze our approach using adversarial attacks, expected calibration error, and visualizations that illustrate the focus of learned representations, to provide a better understanding of our approach. 
We highlight the common denominator between its security vulnerability and decrease in calibration error while illustrating the intended effects of retaining filters that contribute the majority of information between layers.


\section{Related Works}
\label{sec:related_works}
Deep network compression offers a number of strategies to help reduce network size while maintaining high performance, such as low-rank approximations~\cite{Jaderberg_2014,Zhang_2015,Zhang_2016}, quantization~\cite{courbariaux2015binaryconnect,hubara2017quantized,Park_2017,zhou2018adaptive}, knowledge distillation~\cite{Lu_2017,Yim_2017}, and network pruning~\cite{han2015learning,luo2017thinet,lin2019towards}.
In this work, we focus on network pruning since it offers a controlled set up  to study and compare changes in the dynamics of a network when filters are removed.
We broadly classify network pruning methods into two categories, unstructured, and structured, which we describe below.
Also, we highlight common strategies to calculate multivariate dependencies and how they vary from our method.

\subsection{Network Pruning}
\subsubsection{Unstructured pruning}
Some of the earliest in this line of work used the second-order relationship between the objective function and weights of a network to determine which weights to remove or retain~\cite{lecun1990optimal,hassibi1993second}.
Although these methods provide deep insights into the relationships within networks, their dense computational requirements and large runtimes made them less practical.
They were surpassed by an alternative approach that thresholded the weight values themselves to obtain a desired level of sparsity before retraining~\cite{han2015learning}. 
Apart from the simplicity of this approach, it also highlighted the importance of re-training from known minima as opposed to from scratch.
Instead of pruning weights in one shot, \cite{guo2016dynamic} offered a continuous and recursive strategy of using mini-iterations to evaluate the importance of connections using their weights, remove unimportant ones and train the network.
Similarly, \cite{li2016pruning} proposed pruning filters using the $l_1$-norm of their weight values as a measure of importance.
By melding standard network pruning using weights with network quantization and Huffman coding,~\cite{han2015deep} showed superior compression performance compared to any individual pipeline.
However, the direct use of weight values across all these methods does not capture the relationships between different layers or the impact of removing weights on downstream layers.
In \alg, we address this issue by explicitly computing the dependency between filters of successive layers and only retaining filters that contribute a majority of the information.
This ensures that there isn't a severe impact on downstream layers.
 
A subset of methods uses data to derive the importance of filter weights. 
Among them, ThiNet~\cite{luo2017thinet} posed the reconstruction of outcomes with the removal of weights as an optimization objective to decide on weights.
More recently, NISP~\cite{yu2018nisp} used the contribution of neurons towards the reconstruction of outcomes in the second to last layer as a metric to retain/remove filters.
These works represent a shift to data-driven logic to consider the downstream impact of pruning, with the use of external computations (e.g., feature ranking in \cite{yu2018nisp}) combined with the quality of features approximated using weights from a single trial.
Compared to the deterministic relationship between the weights and feature ranking methods, our method uses a probabilistic approach to measure the dependency between filters, thereby accounting for some form of uncertainty.
Further, our method uses one single prune-retrain step compared to multiple iterations of fine-tuning performed in these methods.

 \subsubsection{Structured Pruning}
The shift to structured pruning was based on the idea of seamlessly interfacing with hardware systems as opposed to relying on software accelerators.
One of the first methods to do this extended the original brain damage~\cite{lecun1990optimal} formulation to include fixed pattern masks for specified groups while using a group-sparsity regularizer~\cite{lebedev2016fast}.
This idea was further extended in works that used the group-lasso formulation~\cite{wen2016learning}, individual or multiple $l_n$ norm constraints~\cite{liu2017learning} on the channel parameters and  in works that balanced individual vs. group relationships~\cite{he2017channel,yoon2017combined} to induce sparsity across desired structures. 
These methods explicitly affect the training phase of a chosen network by optimizing a harder and more sensitive objective function.
Further, side-effects like-stability to adversarial attacks, calibration, and difference in learned representations have not been fully quantified.
In our work, we optimize the standard cross-entropy objective function and characterize the behaviour of the original network and their compressed counterparts.

Hybrid methods combine the notion of a modified objective function with the measurement of downstream impact of pruning by enforcing sparsity constraints on the outcomes of groups~\cite{huang2018data}, using a custom group-lasso formulation with a squared dependency on weight values as the importance measure~\cite{li2019oicsr} or an adversarial pruned network generator to compete with the features derived from the original network~\cite{lin2019towards}.
The disadvantages of these methods include their multi-pass scheme and large training times as well as those inherited from modifying the objective function.


\subsection{Multivariate Dependency Measures}
The accurate estimation of multivariate dependency in high-dimensional settings is a hard task. 
The first in this line of work involved Shannon Mutual Information which was succeeded by a number of plug-in estimators including Kernel Density Estimators~\cite{kraskov2004estimating} and KNN estimators~\cite{moon2017ensemble}.
However, their dependence on density estimates and large runtime complexity means they are not suitable for large scale applications including neural networks.
A faster plug-in method based on graph theory and nearest neighbour ratios~\cite{noshad2017direct} was proposed as an alternative.
More solutions that use statistics like KL divergence like ~\cite{leonenko2008class} or Renyi-$\alpha$~\cite{gao2015efficient} were proposed to help bypass density estimation fully.
Instead, in this work, we focus on a conditional GMI estimator, similar to \cite{yasaei2019geometric}, which bypasses the difficult task of density estimation and is non-parametric and scalable.

\section{MINT}
\label{sec:mint}
\alg~is a data-driven approach to pruning networks by learning the dependency between filters/nodes of successive layers.
For every pair of layers in a deep network, \alg~uses the conditional GMI (Section~\ref{subsec:conditional_mutual_information_measure}) between the activations of every filter from a chosen layer $l$ and a filter from layer $l+1$, given the existence of every other possible filter in layer $l$ to compute an importance score.
Here, data flows from layer $l$ to $l+1$.
Once all such importance scores are evaluated, we remove a fixed portion of filters with the lowest scores.
This induces the desired level of sparsity in the network before retraining the network to maintain high accuracy.
The core algorithm is outlined and explained in the sections below.

\subsection{Conditional Geometric Mutual Information}
\label{subsec:conditional_mutual_information_measure}
In this section, we review conditional GMI estimation~\cite{yasaei2019geometric} and use a close approximation to their method to calculate multivariate dependencies in our proposed algorithm.
\hfill \\ \hfill \\
\noindent{\bf{Definition}} 
We first define a general form of GMI denoted by $I_p$: For parameters $p\in(0,1)$ and $q=1-p$ consider two random variables $\BX\in \bbR^{d_x}$ and $\BY\in \bbR^{d_y}$ with joint and marginal distributions $f(\bx,\by)$, $f(\bx)$, and $f(\by)$ respectively. The GMI between $\BX$ and $\BY$ is given by
\begin{multline}
\label{EQ:MIP}  
I_p(\BX;\BY)=\diy\frac{1}{4pq} \times \\ 
\left[\iint \frac{\big( f(\bx,\by)-q f(\by)f(\by)\big)^2}{ pf(\bx,\by)+qf(\bx)f(\by)}\;\rd \bx\; \rd\by -(p-q)^2 \right].
\end{multline}



Considering the special case of $p=q=1/2$ in Eqn.~\ref{EQ:MIP} we obtain,
\begin{equation}
\label{proposed.measure}
I(\BX;\BY)=1-2\diy \iint\frac{ f_{XY}(\bx,\by) f_X(\bx) f_{Y}(\by)}{f_{XY}(\bx,\by) +f_X(\bx)f_{Y}(\by)}\;\rd \bx\;\rd\by.
\end{equation}

The conditional form of this measure is proposed in ~\cite{yasaei2019geometric} as,
\begin{equation}
    I(\BX;\BY|\BZ)=\bbE_Z\left[I(\BX;\BY|\BZ=\bz)\right],\;\;\;\hbox{where}
    \end{equation}
    \begin{multline}
I(\BX;\BY|\BZ=\bz)= \\
1-2\diy\iint \frac{f_{XY|Z}(\bx,\by|\bz) f_{X|Z}(\bx|\bz) f_{Y|Z}(\by|\bz)}{f_{XY|Z}(\bx,\by|\bz)+\;f_{X|Z}(\bx|\bz)f_{Y|Z}(\by|\bz)}\;\rd \bx\;\rd \by. 
\end{multline}

\begin{figure*}[ht!]
    \centering
    \includegraphics[width=\textwidth]{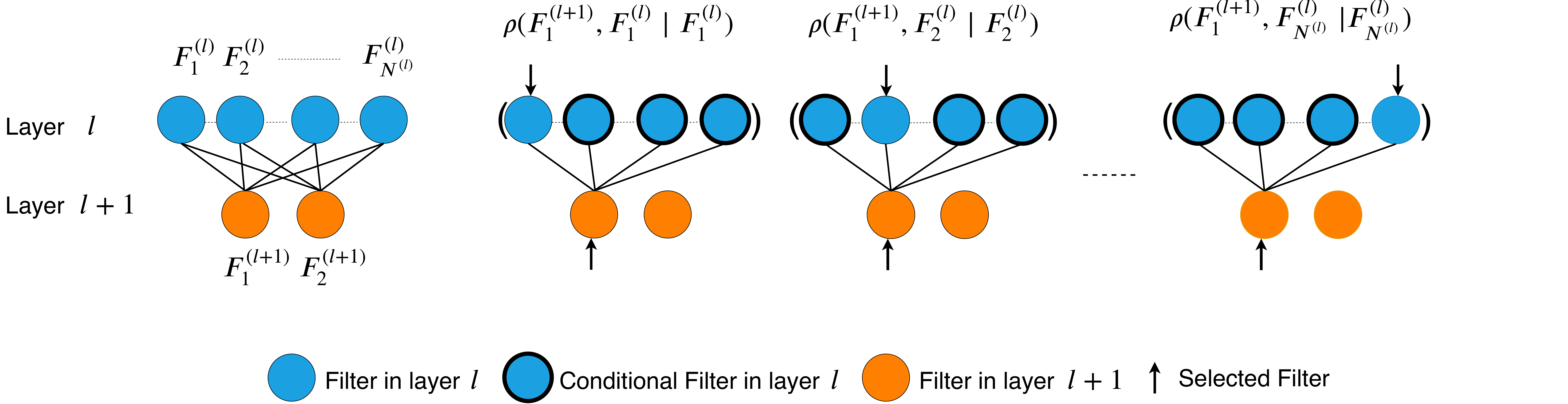}
    \caption{Example of computing multivariate dependencies, $\rho()$, between activations of filter $F^{(L+1)}_1$ and each filter in the previous layer $l$. In each $\rho()$ computation, the highlighted filters in layer $l$ represent the conditioned variables while the arrows indicate the filters whose actual dependence we compute. These steps are repeated for every possible combination of filters across every pair of consecutive layers in the network}
    \label{fig:mint_dependency}
\end{figure*}

\noindent{\bf{Estimator}}
In general, for a set of $m$ samples drawn from $f(\bx,\by,\bz)$, we estimate $I(\BX;\BY|\BZ)$ as follows: (1) split data into two subsets $S_1$ and $S_2$, (2) use the Nearest Neighbour Bootstrap algorithm~\cite{sen2017model} to generate conditionally independent samples from $S_2$ points and name the new set $\bar{S}_2$. (3) Merge $S_1$ and $\bar{S}_2$ i.e. $S:=S_1\bigcup \bar{S}_2$. (4) Construct a Minimum Spanning Tree (MST) on $S$. (5) Compute Friedman-Rafsky statistic \cite{FR}, $\mathfrak{R}_m$, which is the number of edges on the MST linking dichotomous points i.e. edges connecting points in $S_1$ to points in $\bar{S}_2$. (6) The estimate for $I(\BX;\BY|\BZ)$, denoted by $\widehat{I}$, is obtained by $1-\big(\mathfrak{R}_m/m$\big). Note that within the MINT algorithm, we apply the conditional GMI estimator on a sub-sampled set of activations from each filter considered.

\subsection{Approach}
\noindent{\bf{Setup}}
In a deep network containing a total of $L$ layers, we compute the dependency~($\rho$) between filters in every consecutive pair of layers.
Here, the layer $l$ is closer to the input while layer $l+1$ is closer to the output among the chosen pair of consecutive layers.
The activations for a given node in layer $l+1$, are computed as,
\def\bx{\mathbf{x}}
\def\bw{\mathbf{w}}
\def\bb{\mathbf{b}}
\begin{equation}
    F^{(l+1)}(\bx)=\sigma\left({\bw}\bx+\bb\right),
\end{equation}
where $\bx \in \mathbb{R}^{m \times d}$, $m$ is the total number of samples, $d$ is the feature dimension and $\bx$ is the input to a given layer used to compute the activations. $\sigma()$ is an activation function, $\bw \in \mathbb{R}^{N^{(l)}}$, and $\bb$ are the weight vector and bias.

\hfil \\
\noindent{\bf Notations}
\begin{itemize}
    \item $F^{(l+1)}_i$ : The activations from the selected filter $i$ from layer $l+1$.
    \item $N^{(l+1)}$ : Total number of filters in layer $l+1$.
    \item  $S_{F^{(l+1)}_i}$: The set of indices that indicate the values that are retained in the weight vector of the selected filter. 
    \item $\rho{()}$: The dependency between two filters (importance score).
    \item $\overline{ F^{(l)}_j}$: The set of all filters excluding $F^{(l)}_j$.
    \item $\delta$: Threshold on importance score to ensure only strong contributions are retained.
\end{itemize}
\noindent{\bf{Description}}
In every iteration of \alg~(Alg.~\ref{alg:MINT_1b}), we find the set of weight values in $\bw$ to retain while the remaining are zeroed out.

\begin{itemize}
\item For a given pair consecutive of layers $(l, l+1)$, we compute the dependency between every filter in layer $l+1$ in relation to filters in layer $l$. 
The main intent of framing the algorithm in this perspective is that the activations from layers closer to the input have a direct effect on downstream layers while the reverse is not true for a forward pass of the network.

\item Using the activations $F()$ for the selected filters, $(F^{(l+1)}_i, F^{(l)}_j)$ we compute the conditional GMI between them given all the remaining filters in layer $l$, as shown in Fig.~\ref{fig:mint_dependency}. 
This dependency captures the relationship between filters in the context of all the contributions from the preceding layer.
Since the activations of layer $l+1$ are the result of contributions from all filters in the preceding layer, we need to account for this when considering the dependence of activations between two selected filters.

\item Based on the strength of each $\rho{(F^{(l+1)}_i, F^{(l)}_j)}$, the contribution of filters from the previous layer is either retained/removed.
We define a threshold $\delta$ for this purpose, a key hyper-parameter.

\item $S_{F^{(l+1)}_{i}}$ stores the indices of all filters from layer $l$ that are retained for a selected filter $F^{(l+1)}_{i}$.
The weights for retained filters are left the same while the weights for the entire kernel in the other filters are zeroed out. 
In the context of fully connected layers, we retain or zero out specific weight values in the weight matrix.

\end{itemize}

\begin{algorithm}[t]
\SetAlgoLined
 \For{Every pair of layers $(l, l+1), l \in {1,2,\dots,L-1}$}
 {
 \For{$F^{(l+1)}_i, i \in 1,2,\dots N^{(l+1)}$}
 {
 Initialize $S_{F^{(l+1)}_i} = \emptyset$\;
 \For{$F^{(l)}_j, j \in 1,2,\dots N^{(l)}$}
 {
 $\rho{(F^{(l+1)}_i, F^{(l)}_j)}$ = $I(F^{(l+1)}_i, F^{(l)}_j\mid \overline{ F^{(l)}_j})$\;
 \If{$\rho{(F^{(l+1)}_i, F^{(l)}_j)} \geq \delta$}
 {
 $S_{F^{(l+1)}_i} = S_{F^{(l+1)}_i} \cup F^{(l)}_j$
 }
 }
 }
 }
\caption{{\alg}  pruning between filters of layers ($l, l+1$)}
\label{alg:MINT_1b}
\end{algorithm}

\noindent{\bf{Group Extension}}
While evaluating dependencies between every pair of filters allows  us to take a close look at their relationships, it does not scale well to deeper or wider architectures.
To address this issue, we evaluate filters in groups rather than one-by-one.
We define $G$ as the total number of groups in a layer, where each group contains an equal number of filters.
We explore in detail the impact of varying the number of groups in Section~\ref{sec:hyper_parameter_empirical_analysis}. 
Although there are multiple approaches to grouping filters, in this work we restrict ourselves to sequential grouping, where groups are constructed from consecutive filters.
There is no explicit requirement for a pre-grouping step before our algorithm so long as a balanced grouping of filters is used.
\hfill \\

\noindent{\bf{Finer Details}}
\alg~is constructed on the assumption that the majority of information from the preceding layer is available and the filter in consideration can selectively retain contributions for a subset of previous filters.
This allows us to work on isolated pairs of layers with minimal interference on downstream layers since retaining filters with high MI will ensure the retention of filters that contribute the most information to the next layer.
By maintaining as much information as possible between layers, the amount of critical information passed to layers further on is maintained.

\section{Experimental Results}
\label{sec:experimental_results}
We breakdown the experimental results into three major sections. Section~\ref{sec:comparison_against_existing_methods} focuses on the comparison of our method to state-of-the-art deep network compression algorithms, Section~\ref{sec:hyper_parameter_empirical_analysis} highlights the significance of various hyper-parameters used in \alg, and finally, Section~\ref{sec:characterization} characterizes \alg-compressed networks w.r.t. their response to adversarial attacks, calibration statistics and learned representations in comparison to their original counterparts. 
As a prelude to all these three parts, we describe the datasets, models, and metrics used across our experiments.
We restrict the implementation details to the appendices.

\subsection{Datasets and Models}
Our experiments are divided into the following Dataset-Architecture combinations in order of increasing complexity in dataset and architectures, MNIST~\cite{lecun1998gradient} + Multi-Layer Perceptron~\cite{almeida1997c1}, CIFAR10~\cite{krizhevsky2009learning} + VGG16~\cite{simonyan2014very}, CIFAR10 + ResNet56 \cite{He_2016} and finally ILSVRC2012~\cite{ILSVRC15} + ResNet50.


\subsection{Metrics}
The key metrics we use to evaluate the performance of various methods are,
\begin{itemize}
    \item Parameters Pruned (\%): The ratio of parameters removed to the total number of parameters in the trained baseline network. A higher value in conjunction with good performance indicates a superior method.
    \item Performance (\%): This performance indicates the best testing accuracy upon training, for baseline networks, and re-training, for pruning methods. A value closer to the baseline performance is preferred.
    \item Memory Footprint (Mb): The amount of memory consumed when storing the weights of a network in PyTorch's~\cite{NEURIPS2019_9015} sparse matrix format (CSR). We compute this value by using SciPy~\cite{2020SciPy-NMeth} to convert weight matrices to CSR format and storing them in ``npz'' files.
\end{itemize}

\begin{figure}[t!]
    \centering
    \subfloat[\label{fig:groups}]{\includegraphics[width=0.45\columnwidth]{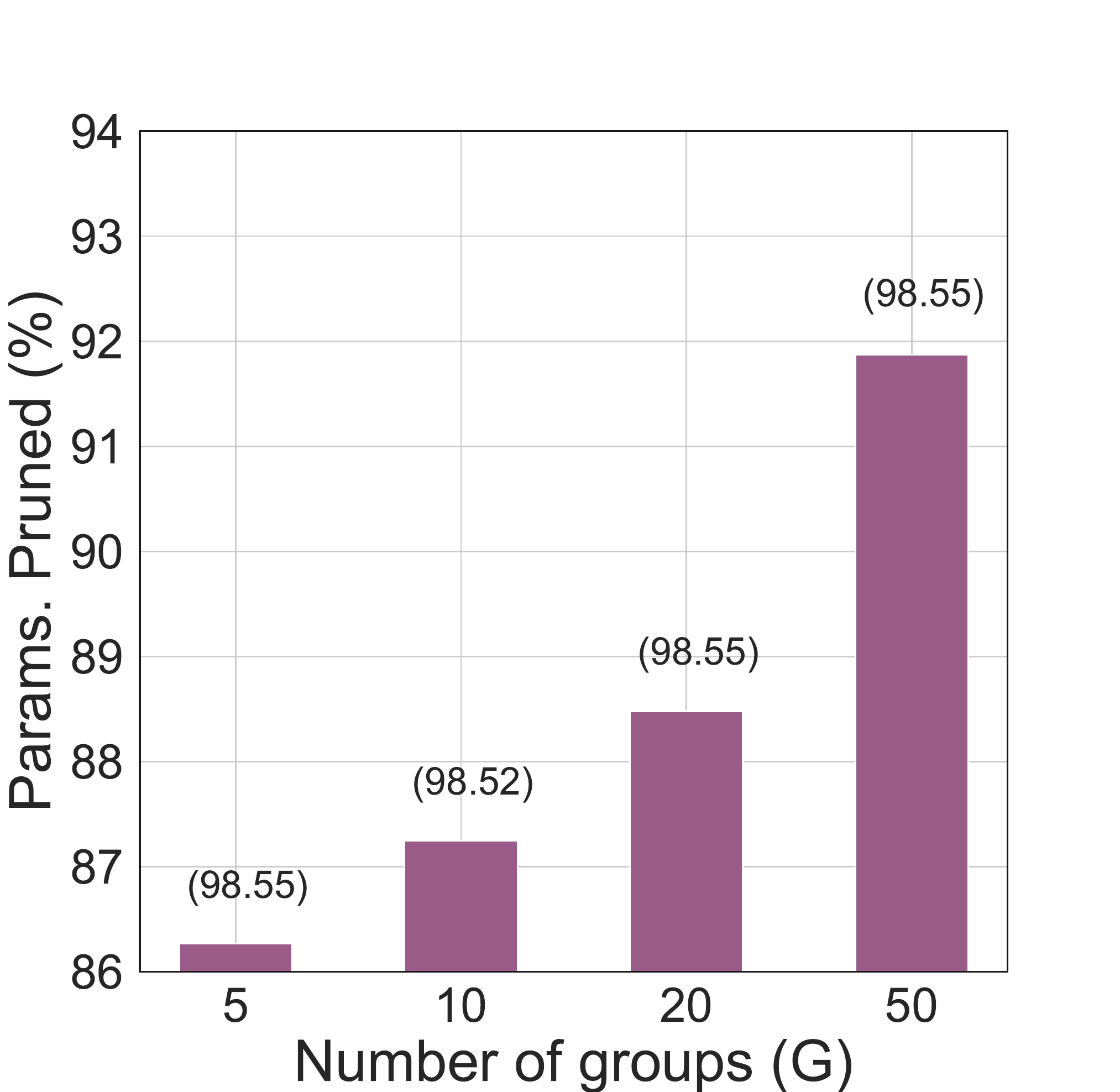}}
    \subfloat[\label{fig:samples}]{\includegraphics[width=0.45\columnwidth]{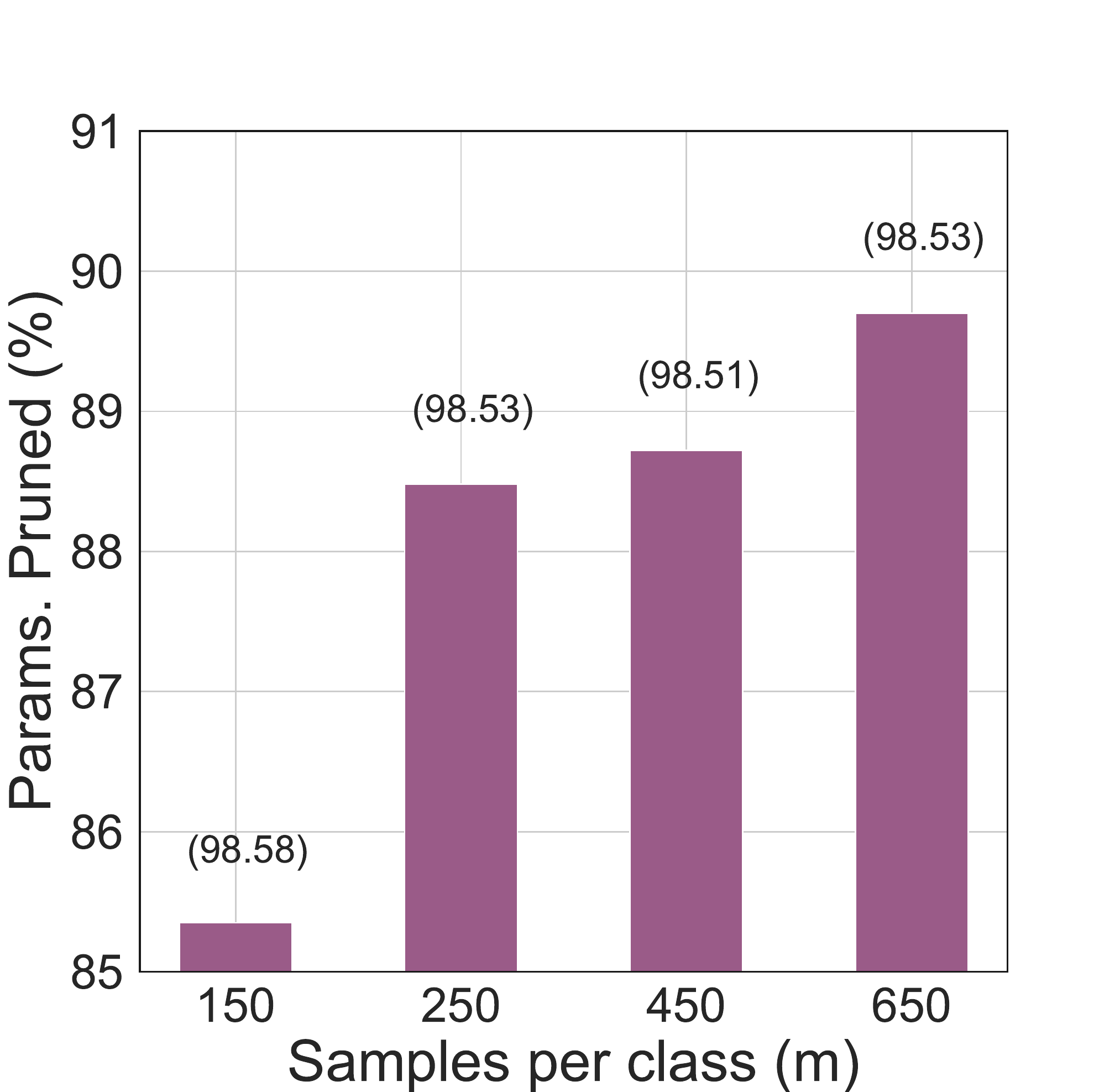}}
    \caption{(a) An increase in the number of groups per layer allows for finer grouping of filters which in turn leads to more accurate GMI estimates and thresholding. Thus, there is a steady increase in the number of parameters that can be removed to achieve $> 98.50$ performance. (b) Keeping $G=20$, we observe that increasing the number of samples per class improves the GMI estimate accuracy which in turn allows for better thresholding and an increase in the parameters pruned. The values on top of the bar plots are the recognition accuracies}
    \label{fig:hyper_parameters_g_samples}
\end{figure}

\begin{table*}[t]
\caption{\alg~is easily able to match (single-step on ILSVRC2012) or outperform (remaining datasets) SOTA pruning methods across the evaluated benchmarks, using only a single prune-retrain step. Baselines are arranged in increasing order of Parameters Pruned $\%$. $^*$ indicates comparison of layer 2's weights}
\label{tab:results_main}
\centering
\begin{tabular}{llccc}
\toprule
                              & Method                   & Params. Pruned($\%$)& Performance($\%$) & Memory(Mb)  \\\midrule
\multirow{5}{*}{\shortstack[l]{MLP \\ MNIST}} & Baseline                & N.A.                    & 98.59           & 0.539 \\ 
                              & SSL~\cite{wen2016learning}              & 90.95$^*$              & 98.47           & N.A. \\
                              & Network Slimming~\cite{liu2017learning} & 96.00$^*$              & 98.51           & N.A. \\ \cline{2-5}
                              & \textbf{MINT} (\textbf{ours})($\delta=0.645$)     & 96.01$^*$     & 98.47           & 0.025 \\ 
                              \midrule
\multirow{5}{*}{\shortstack[l]{VGG16 \\CIFAR-10}} & Baseline & N.A.         & 93.98                                         & 53.904 \\ 
                              & Pruning Filters~\cite{li2016pruning}  & 64.00       & 93.40           & N.A \\
                              & SSS~\cite{huang2018data}              & 73.80       & 93.02           & N.A \\
                              & GAL~\cite{lin2019towards}             & 82.20       & 93.42           & N.A. \\ \cline{2-5}
                              & \textbf{MINT} (\textbf{ours}) ($\delta=0.850$)     & 83.43     & 93.43           & 9.057 \\ \midrule
\multirow{7}{*}{\shortstack[l]{ResNet56 \\ CIFAR-10}}  & Baseline & N.A.     & 92.55                          & 3.110 \\ 
                              & GAL~\cite{lin2019towards}             & 11.80              & 93.38           & N.A. \\
                              & Pruning Filters~\cite{li2016pruning}  & 13.70              & 93.06           & N.A. \\
                              & NISP~\cite{yu2018nisp}                & 42.40              & 93.01           & N.A. \\
                              & OED~\cite{wen2016learning}            & 43.50              & 93.29           & N.A. \\ \cline{2-5}
                              & \textbf{MINT} (\textbf{ours}) ($\delta=0.184$)     & 52.41     & 93.47  & 1.553 \\ 
                              & \textbf{MINT} (\textbf{ours}) ($\delta=0.208$)     & 55.39     & 93.02  & 1.462 \\ \midrule
\multirow{8}{*}{\shortstack[l]{ResNet50 \\ ILSVRC2012}}  & Baseline & N.A.     & 76.13                                         & 91.163 \\ 
                              & GAL~\cite{lin2019towards}              & 16.86              & 71.95           & N.A. \\
                              & OED~\cite{wen2016learning}             & 25.68              & 73.55           & N.A. \\
                              & SSS~\cite{huang2018data}               & 27.05              & 74.18           & N.A. \\
                              & ThiNet~\cite{luo2017thinet}            & 51.45              & 71.01           & N.A. \\ \cline{2-5}
                              & \textbf{MINT} (\textbf{ours}) ($\delta=0.1000$)     & 43.00     & 71.50           & 52.371 \\ 
                              & \textbf{MINT} (\textbf{ours}) ($\delta=0.1101$)     & 49.00     & 71.12           & 47.519 \\ 
                              & \textbf{MINT} (\textbf{ours}) ($\delta=0.1103$)     & 49.62     & 71.05           & 46.931 \\ 
                            \bottomrule
\end{tabular}
\end{table*}

\begin{figure*}[h!]
    \centering
    \subfloat[\label{fig:sens_vgg16}]{\includegraphics[width=0.33\textwidth]{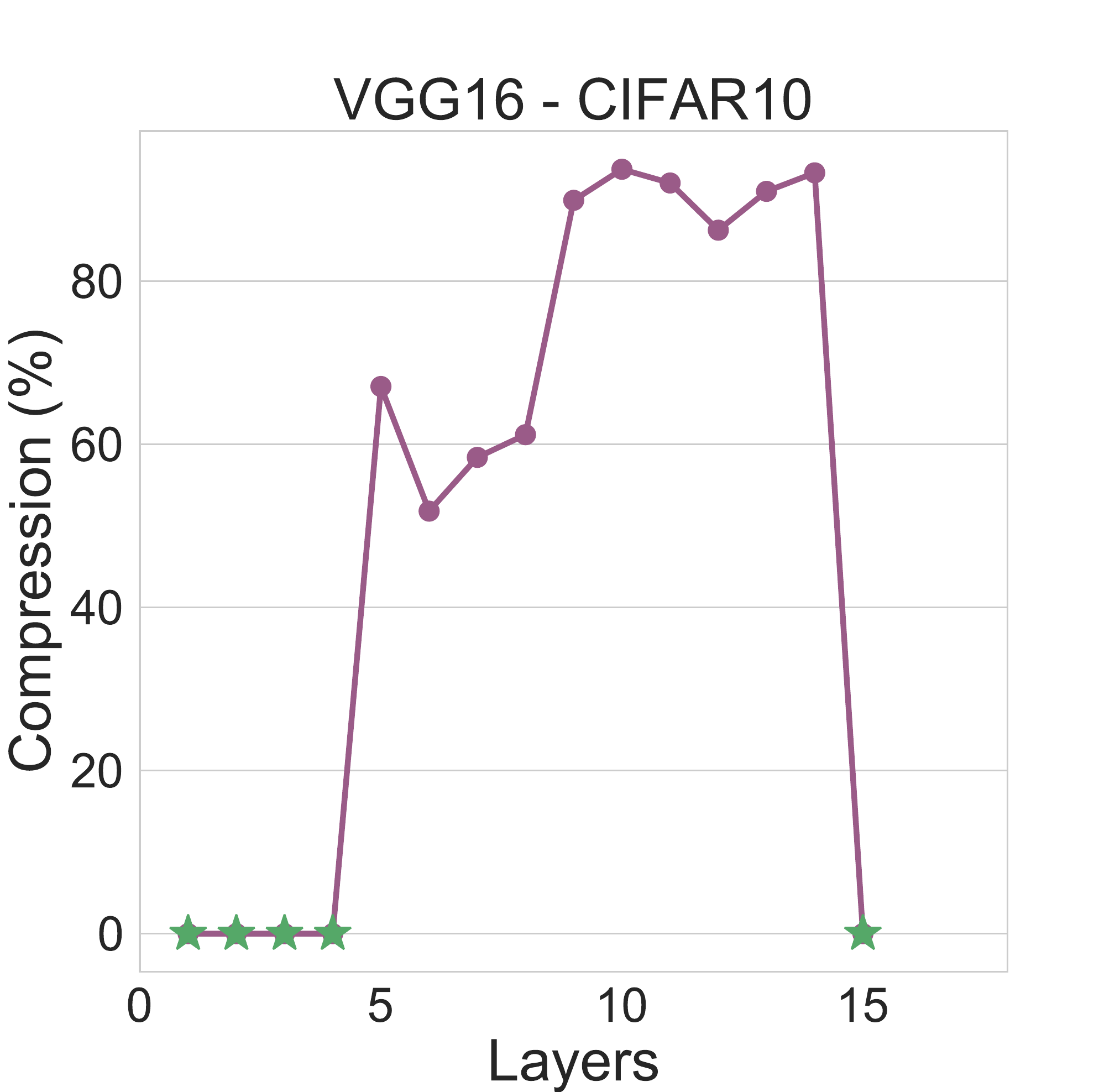}}
    \subfloat[\label{fig:sens_resnet56}]{\includegraphics[width=0.33\textwidth]{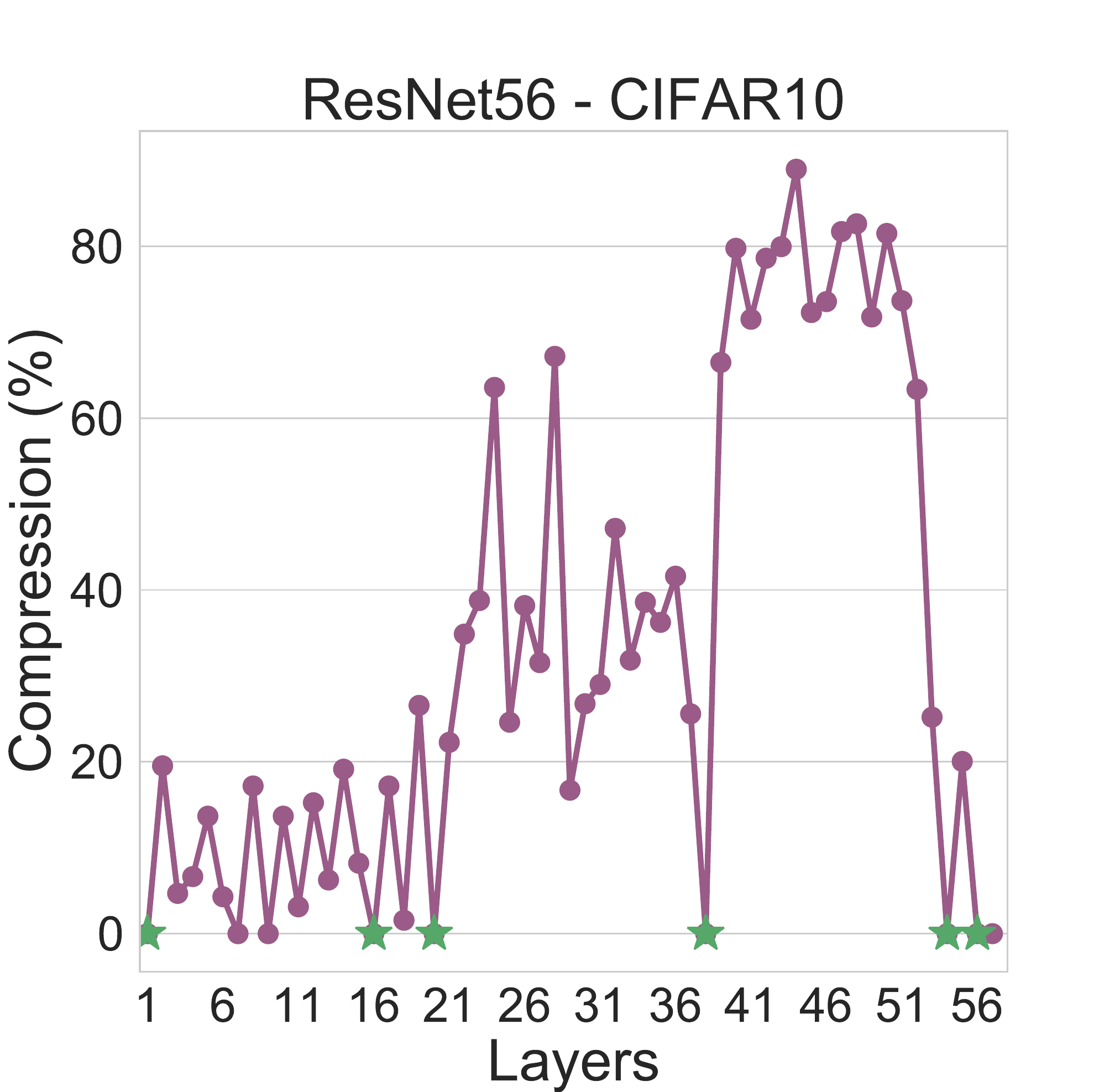}}
    \subfloat[\label{fig:sens_resnet50}]{\includegraphics[width=0.33\textwidth]{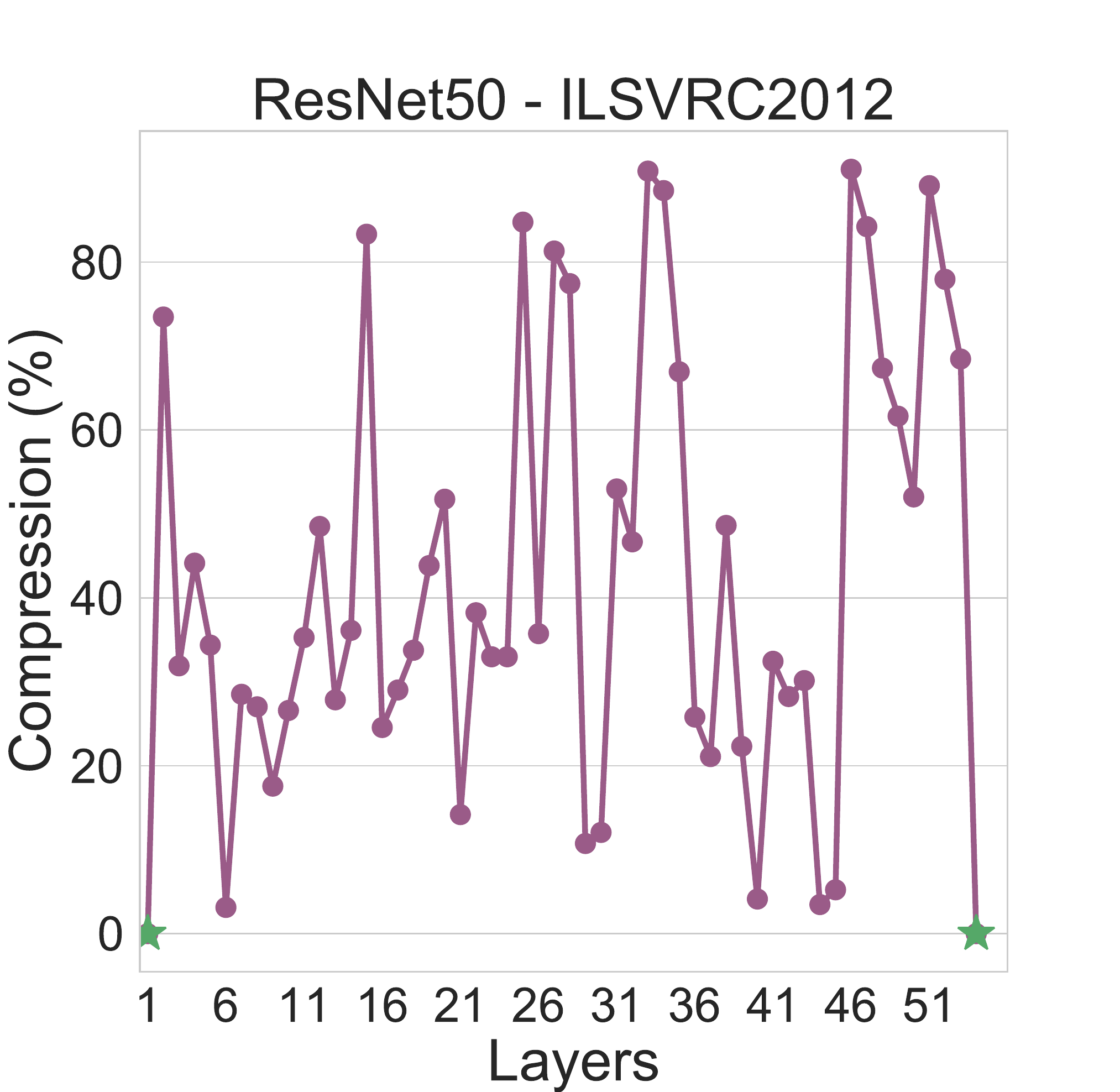}}
    \caption{Illustration of compression $\%$ per layer for the smallest \alg-compressed networks. We observe a characteristic peak in compression towards the later layers of both VGG16 and ResNet56 when trained on CIFAR-10. However, compression is spread over the course the entire network for ResNet50.
    The green stars indicate layers we avoid pruning due to high sensitivity}
    \label{fig:sens_per_layer}
\end{figure*}

\begin{figure*}[ht!]
    \centering
    \subfloat[\label{fig:adv_mlp}]{\includegraphics[width=0.25\textwidth]{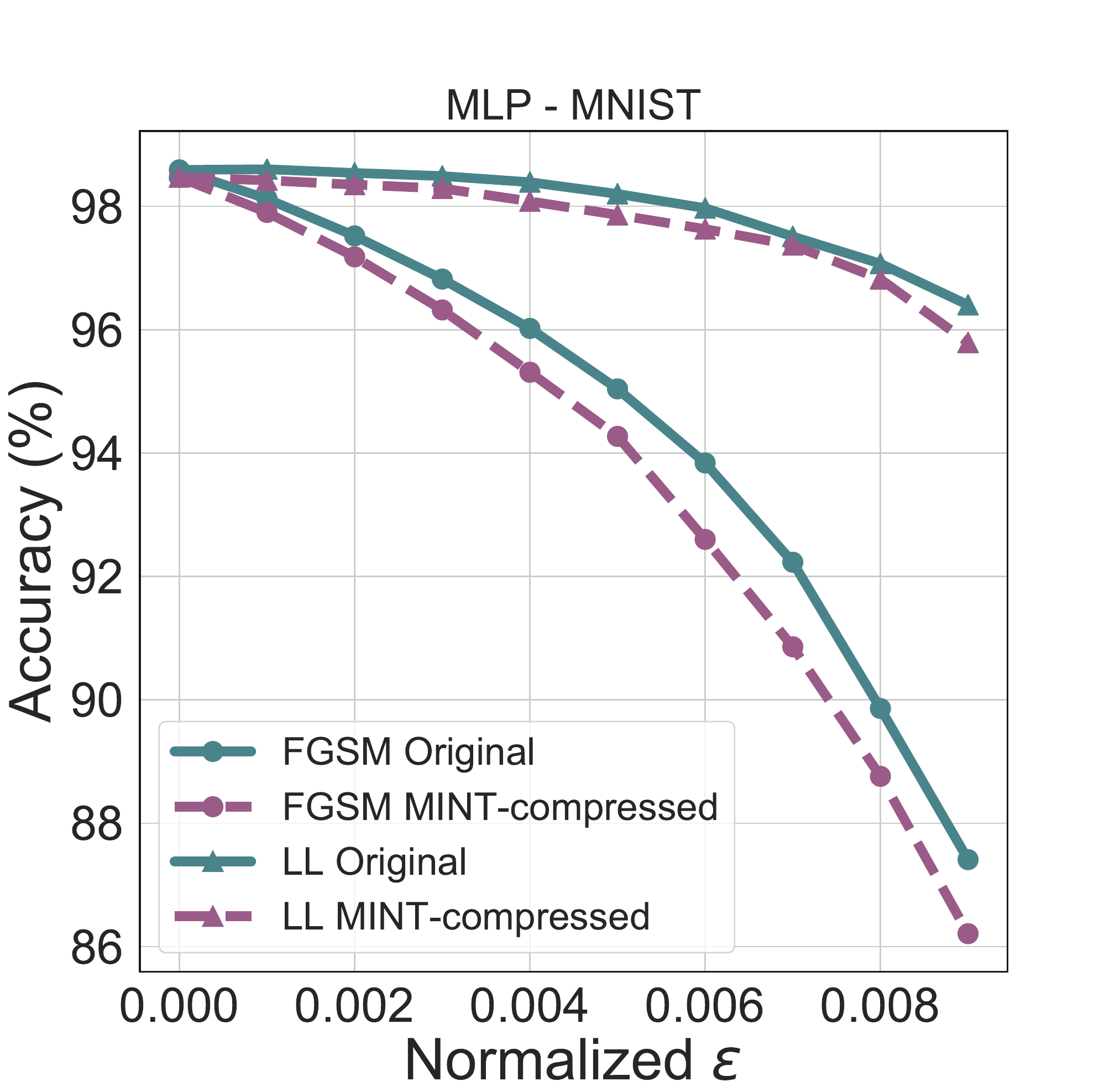}}
    \subfloat[\label{fig:adv_vgg16}]{\includegraphics[width=0.25\textwidth]{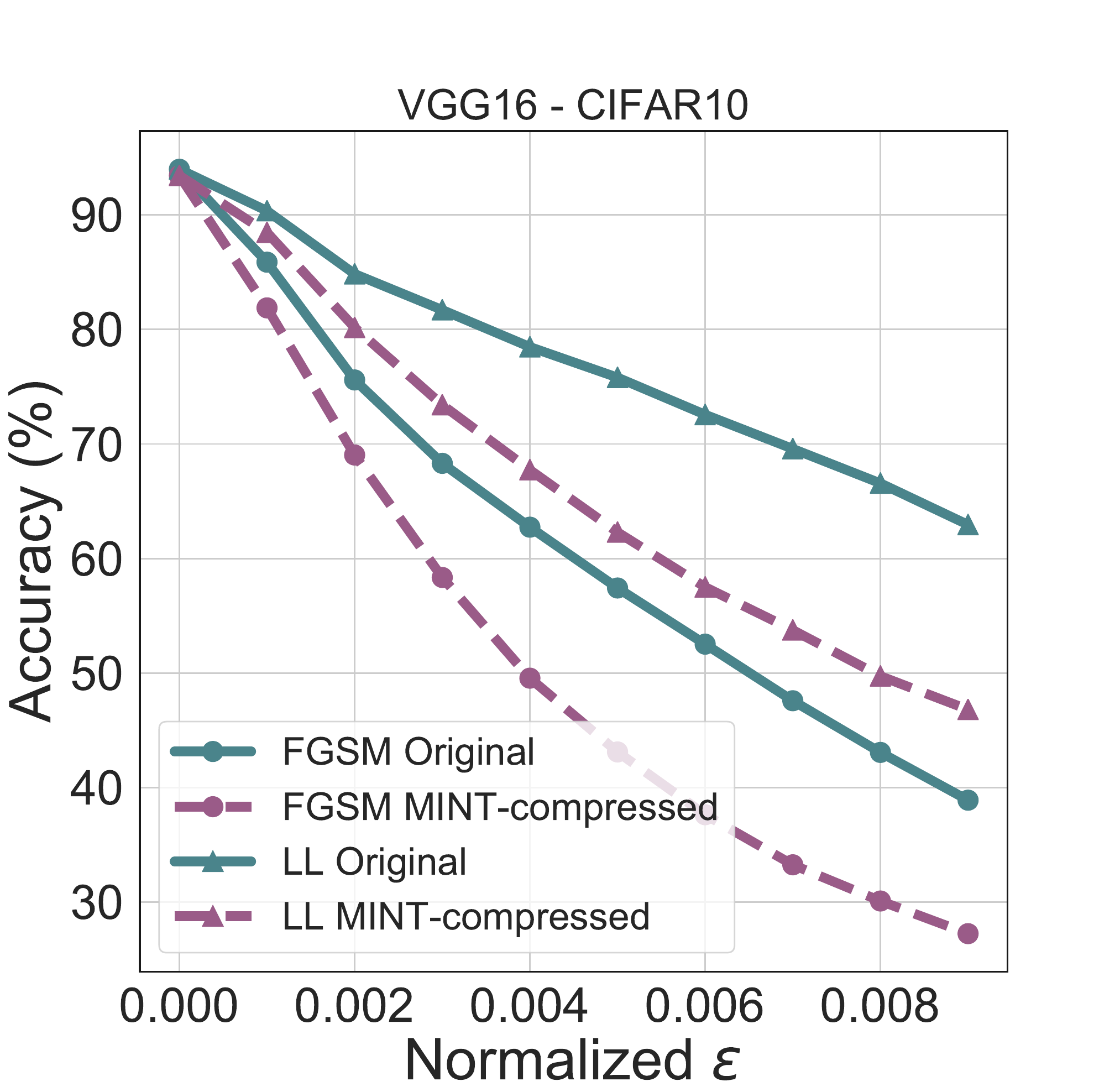}}
    \subfloat[\label{fig:adv_resnet56}]{\includegraphics[width=0.25\textwidth]{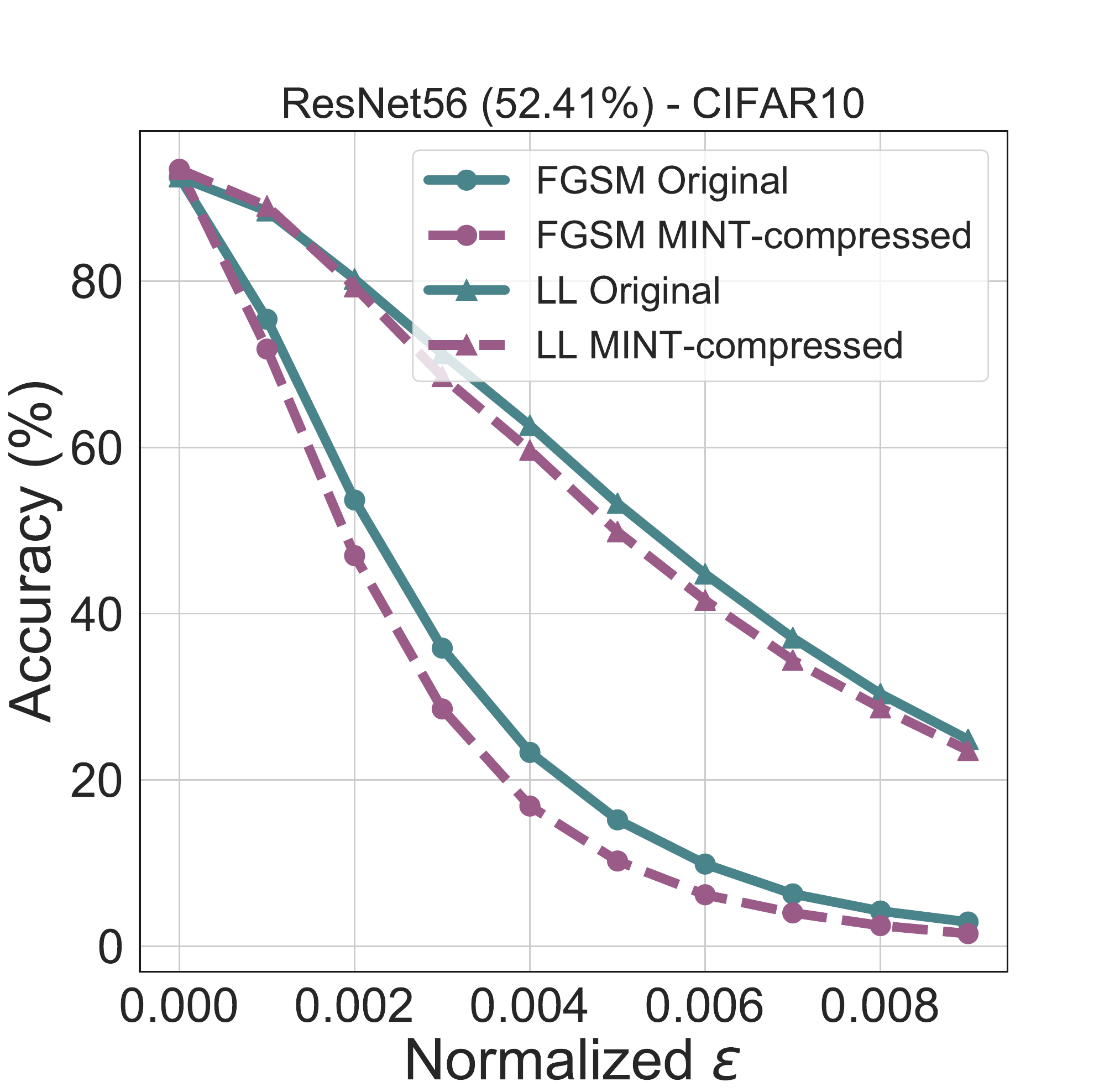}}
    \subfloat[\label{fig:adv_resnet50}]{\includegraphics[width=0.25\textwidth]{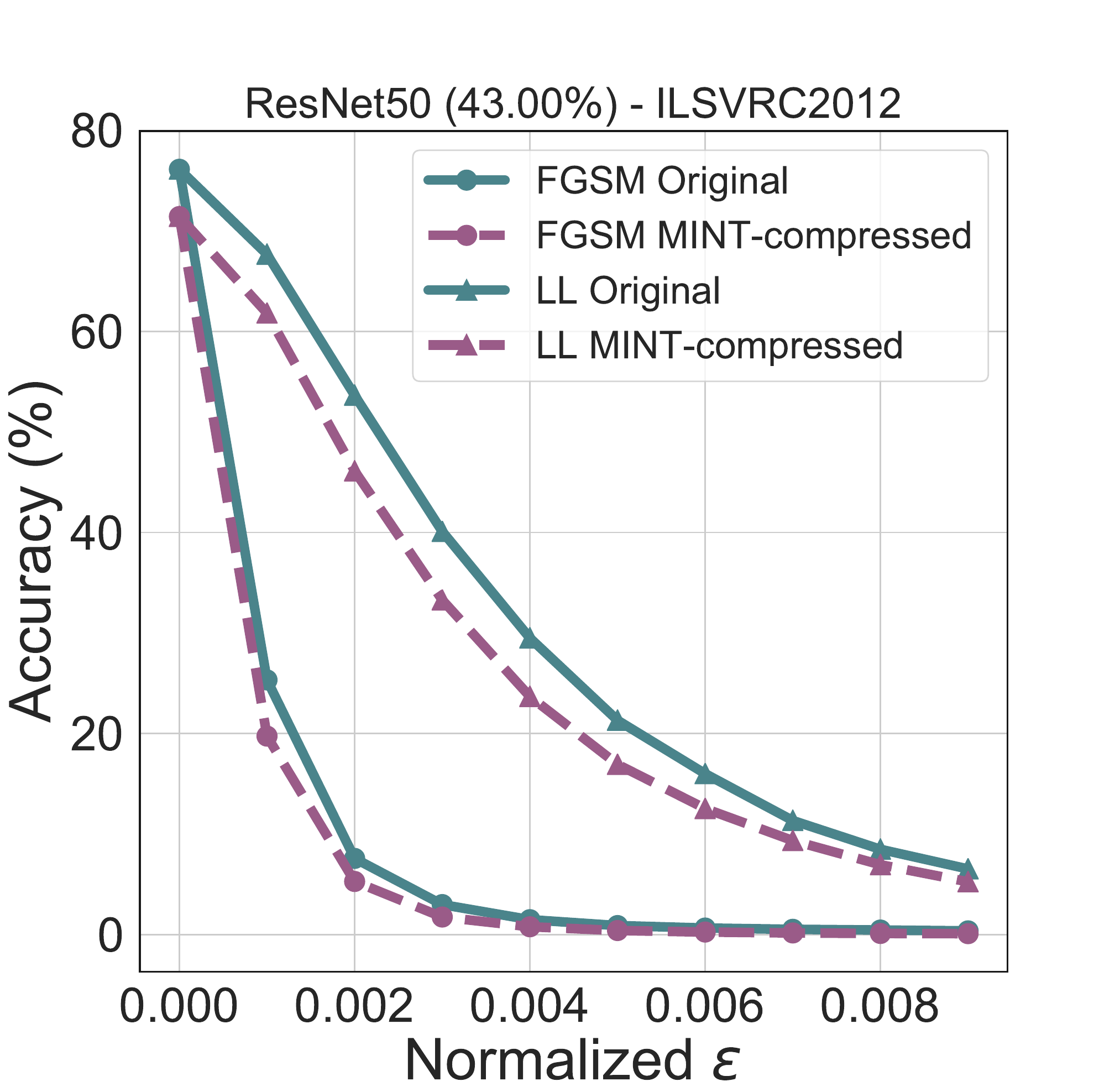}}
    \caption{By enforcing the use of an important subset of filters from all the available ones, \alg-compressed networks begin to overvalue their importance. By emphasizing a small set of features, it makes \alg-compressed networks more susceptible to both targeted and non-targeted adversarial attacks when compared to the original network. Here, $\epsilon$ refers to the $\epsilon$ ball in $l_\infty$ norm}
    \label{fig:adversarial_attacks}
\end{figure*}

\begin{figure*}[ht!]
    \centering
    \includegraphics[width=0.6\textwidth]{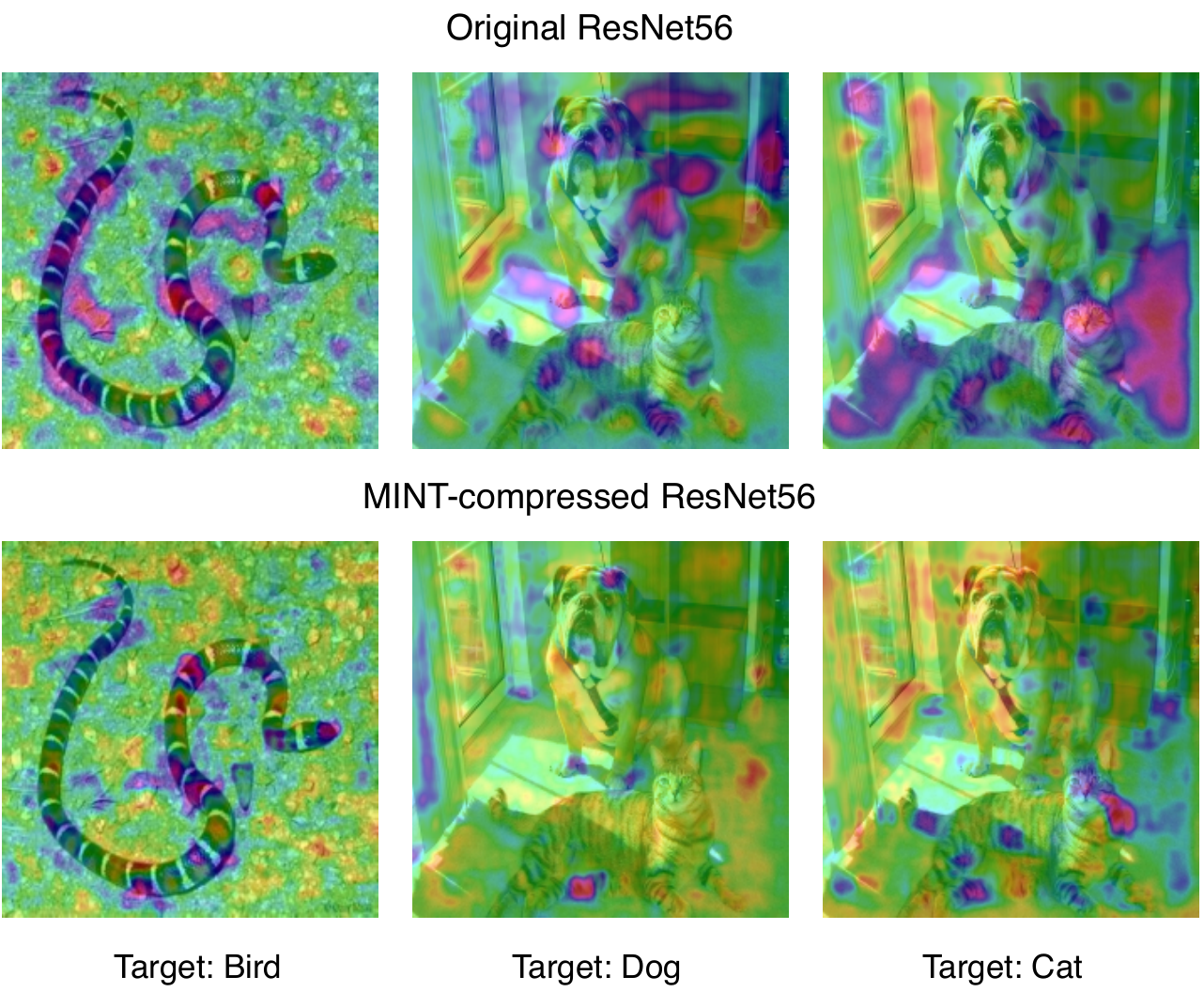}
    \caption{Visualizations using GradCAM~\cite{selvaraju2017grad} illustrate the decrease in effective portions of the image that contribute towards the target decisions in \alg-compressed ResNet56 (row 2) when compared to the original network (row 1)}
    \label{fig:learned_representations}
\end{figure*}

\begin{figure*}[ht!]
    \centering
    \subfloat[ECE:0.0077]{\label{fig:cal_mlp}\includegraphics[width=0.25\textwidth]{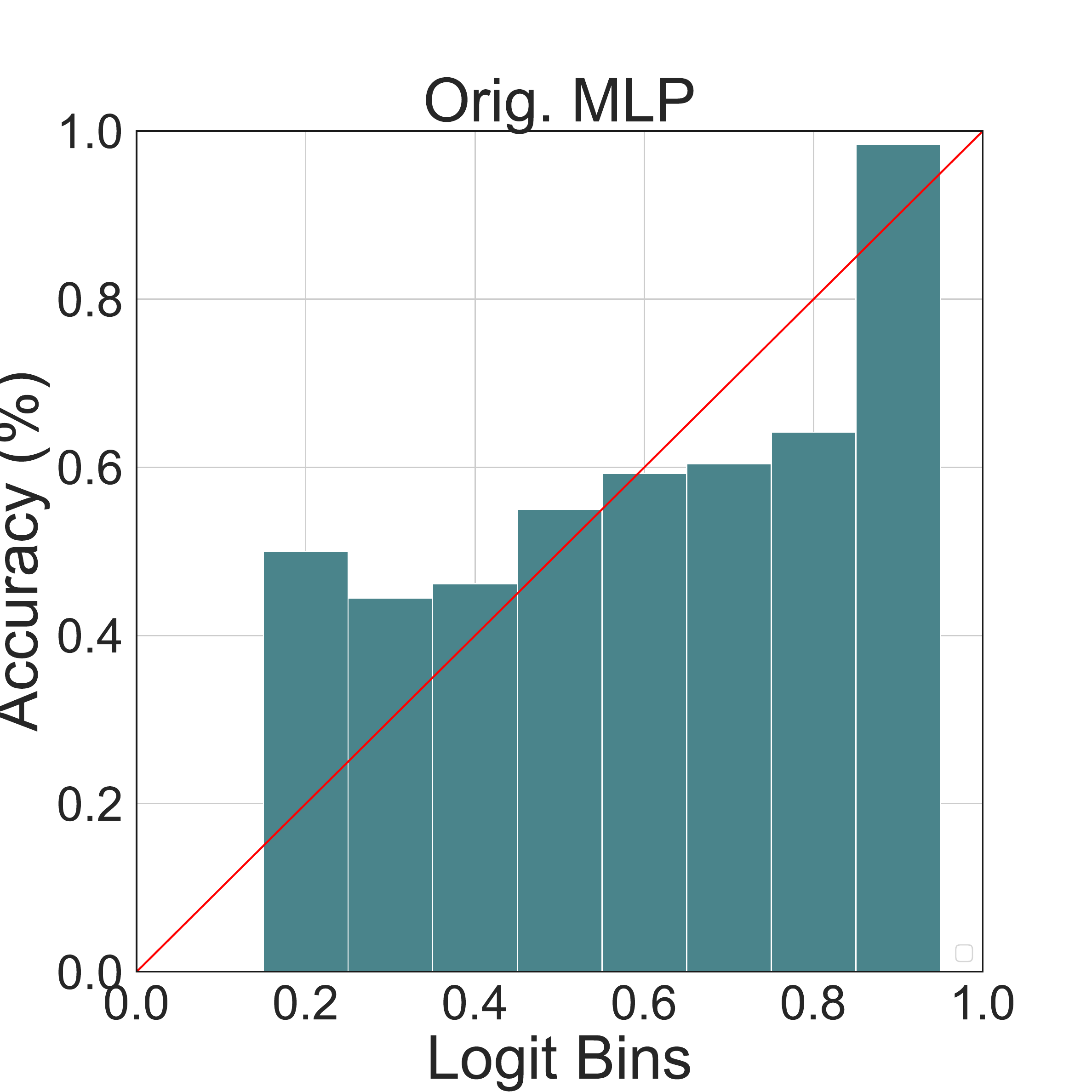}}
    \subfloat[ECE:0.0517]{\label{fig:cal_vgg16}\includegraphics[width=0.25\textwidth]{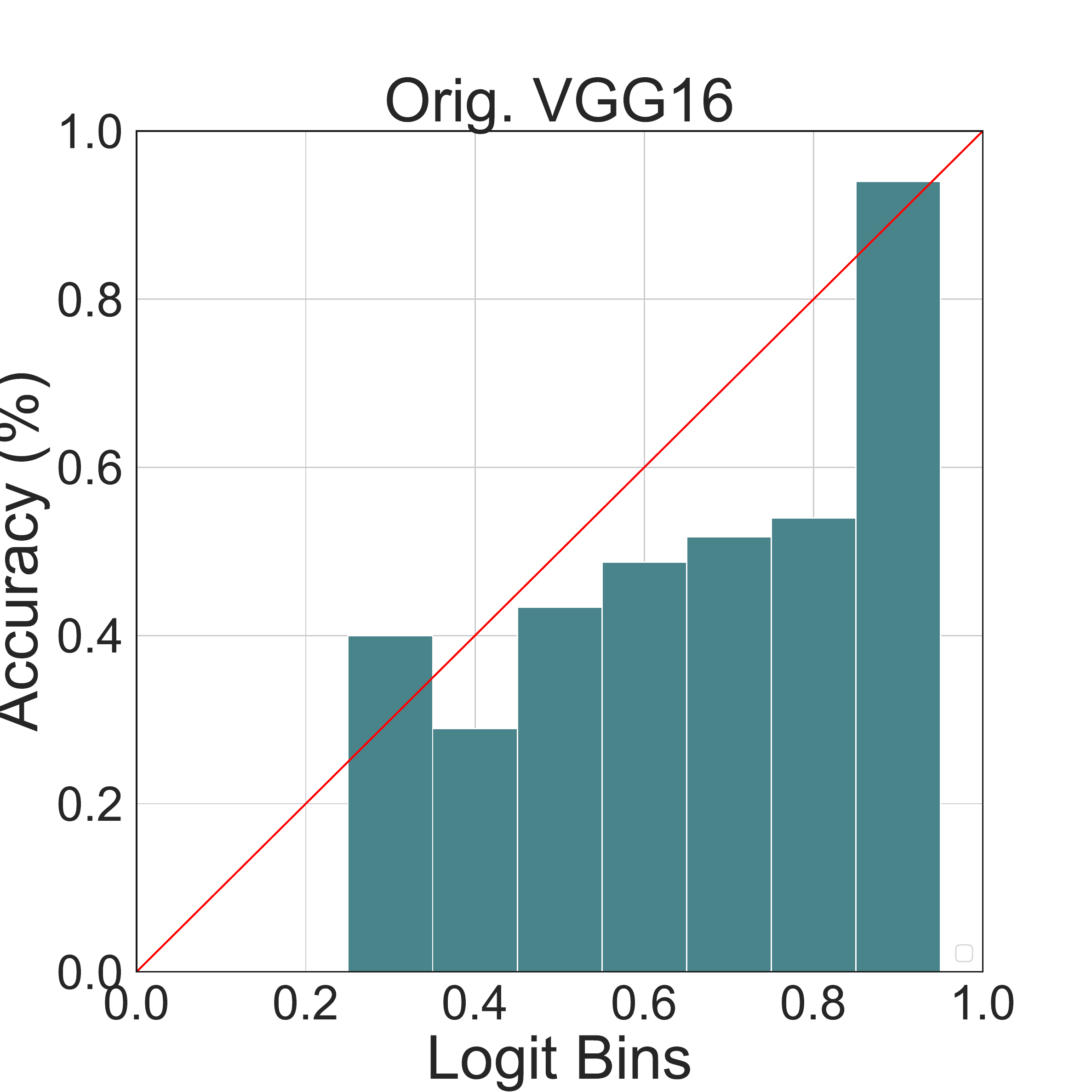}}
    \subfloat[ECE:0.0762]{\label{fig:cal_resnet56}\includegraphics[width=0.25\textwidth]{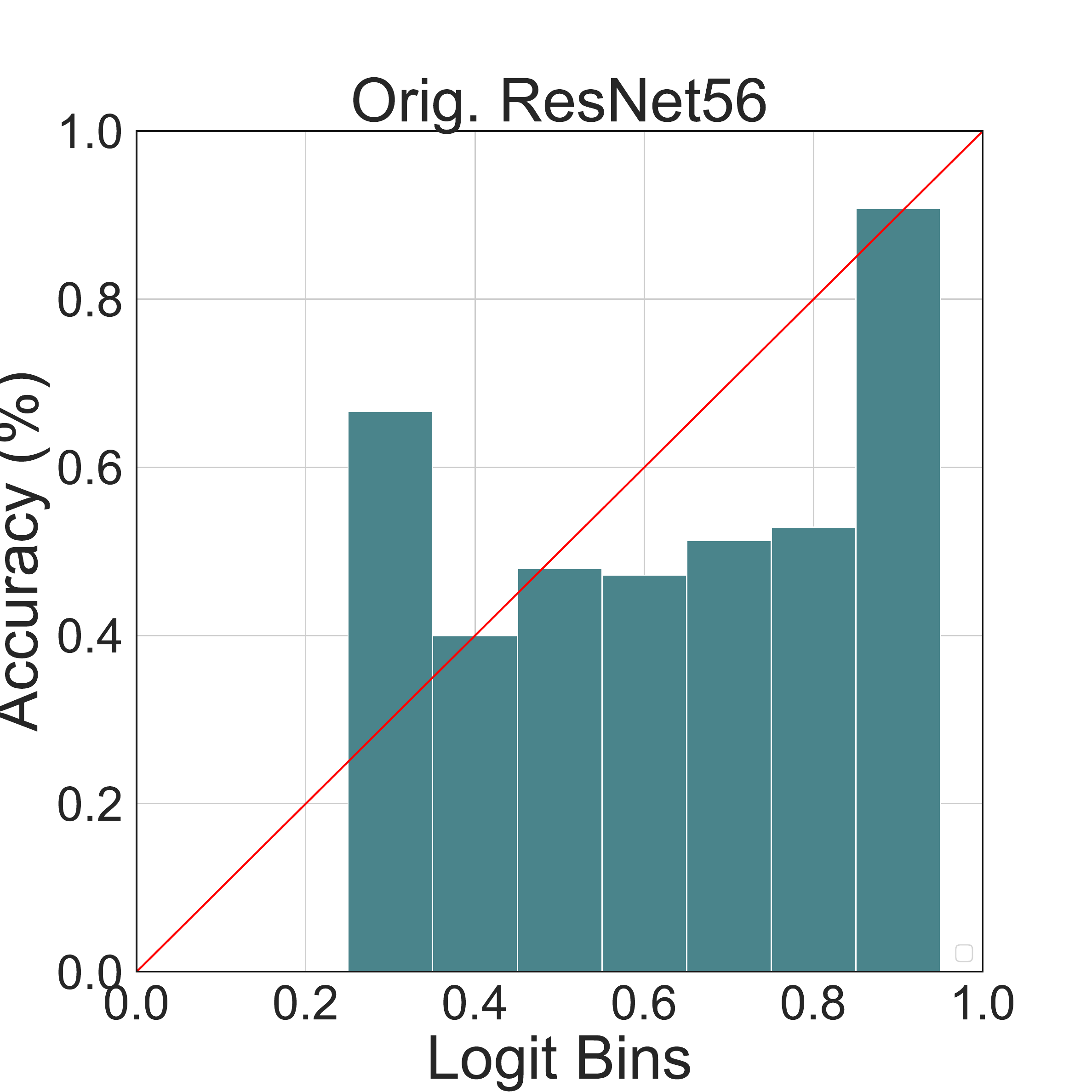}}
    \subfloat[ECE:0.0305]{\label{fig:cal_resnet50}\includegraphics[width=0.25\textwidth]{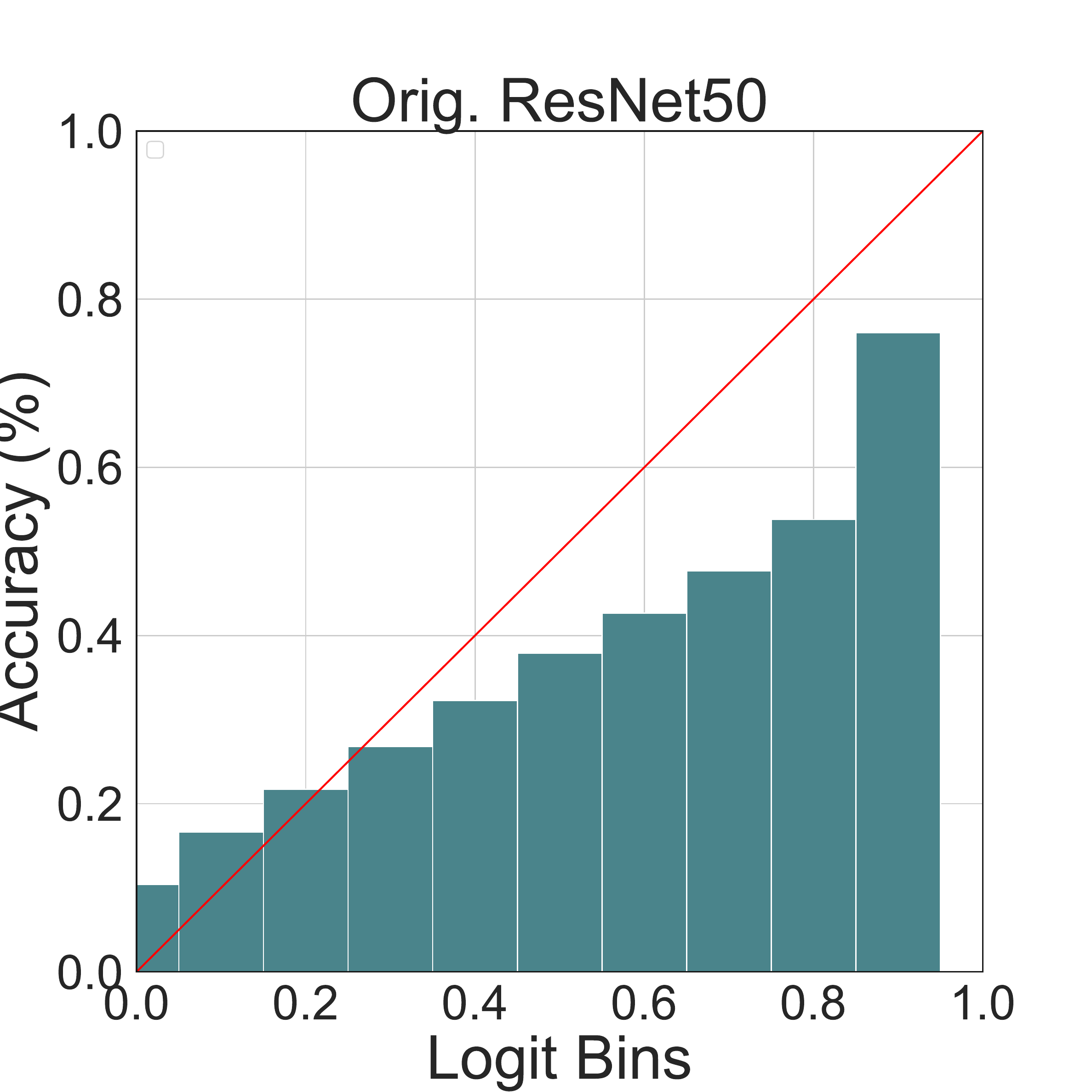}}
    
    \subfloat[ECE:0.0054]{\label{fig:cal_mint_mlp}\includegraphics[width=0.25\textwidth]{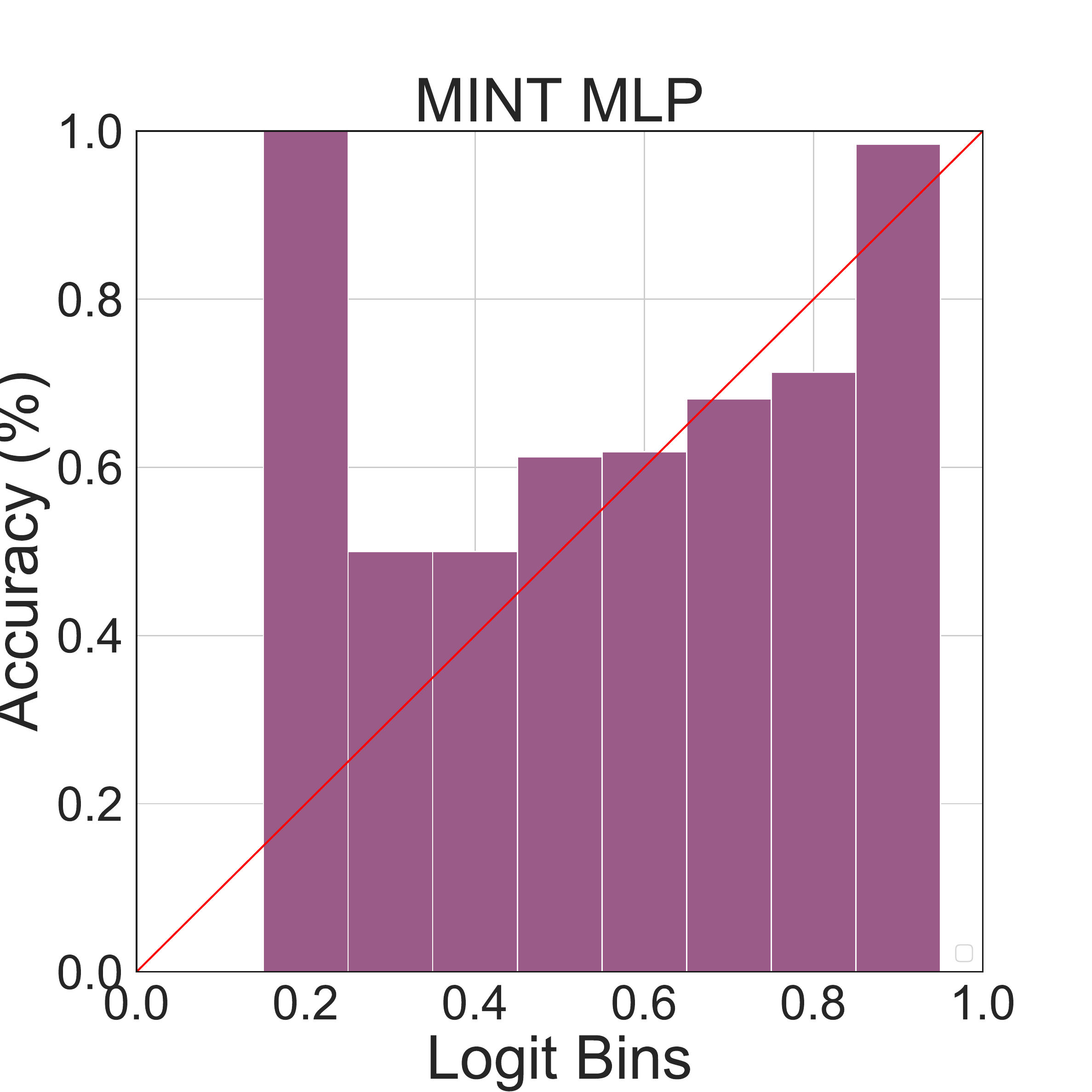}}
    \subfloat[ECE:0.0500]{\label{fig:cal_mint_vgg16}\includegraphics[width=0.25\textwidth]{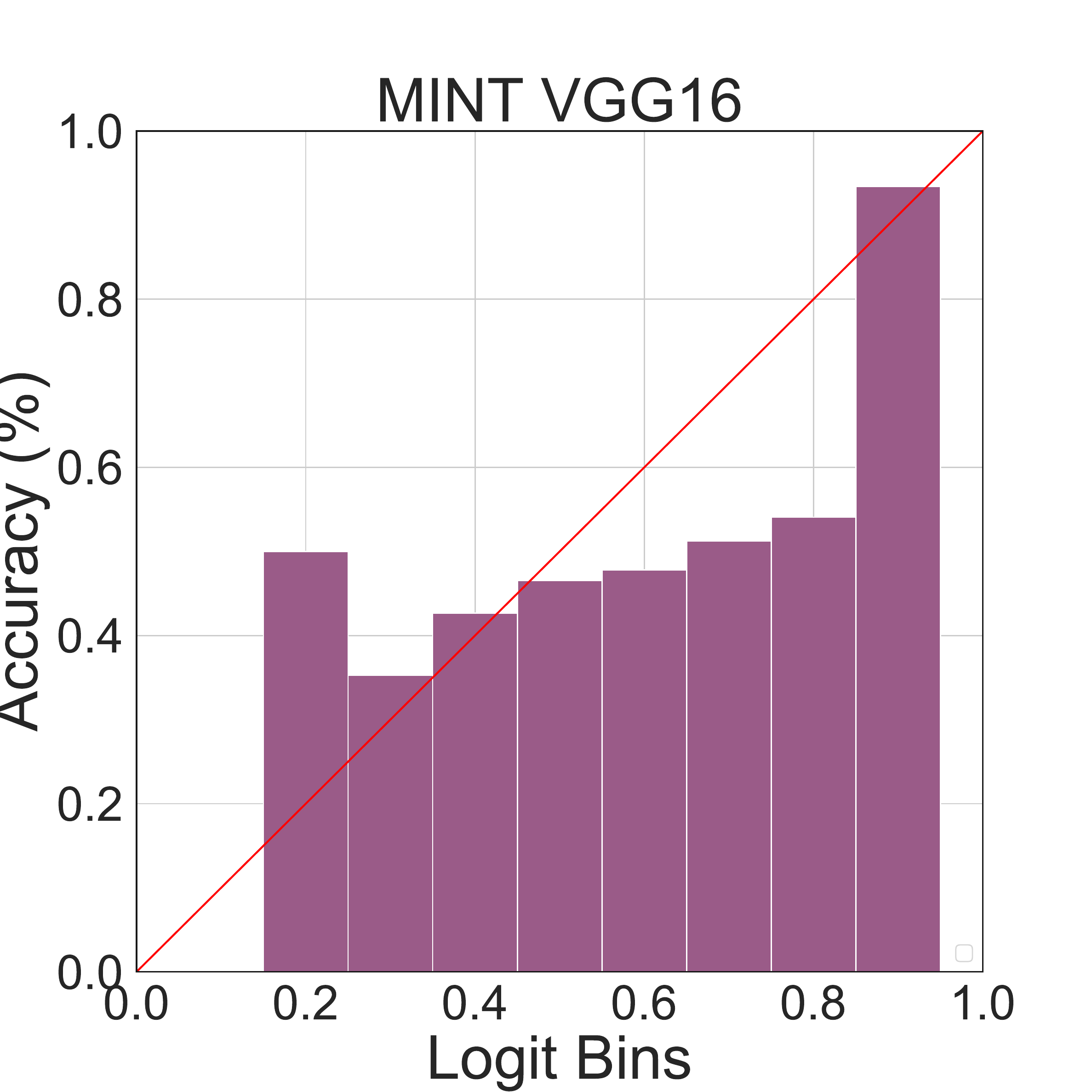}}
    \subfloat[ECE:0.0383]{\label{fig:cal_mint_resnet56}\includegraphics[width=0.25\textwidth]{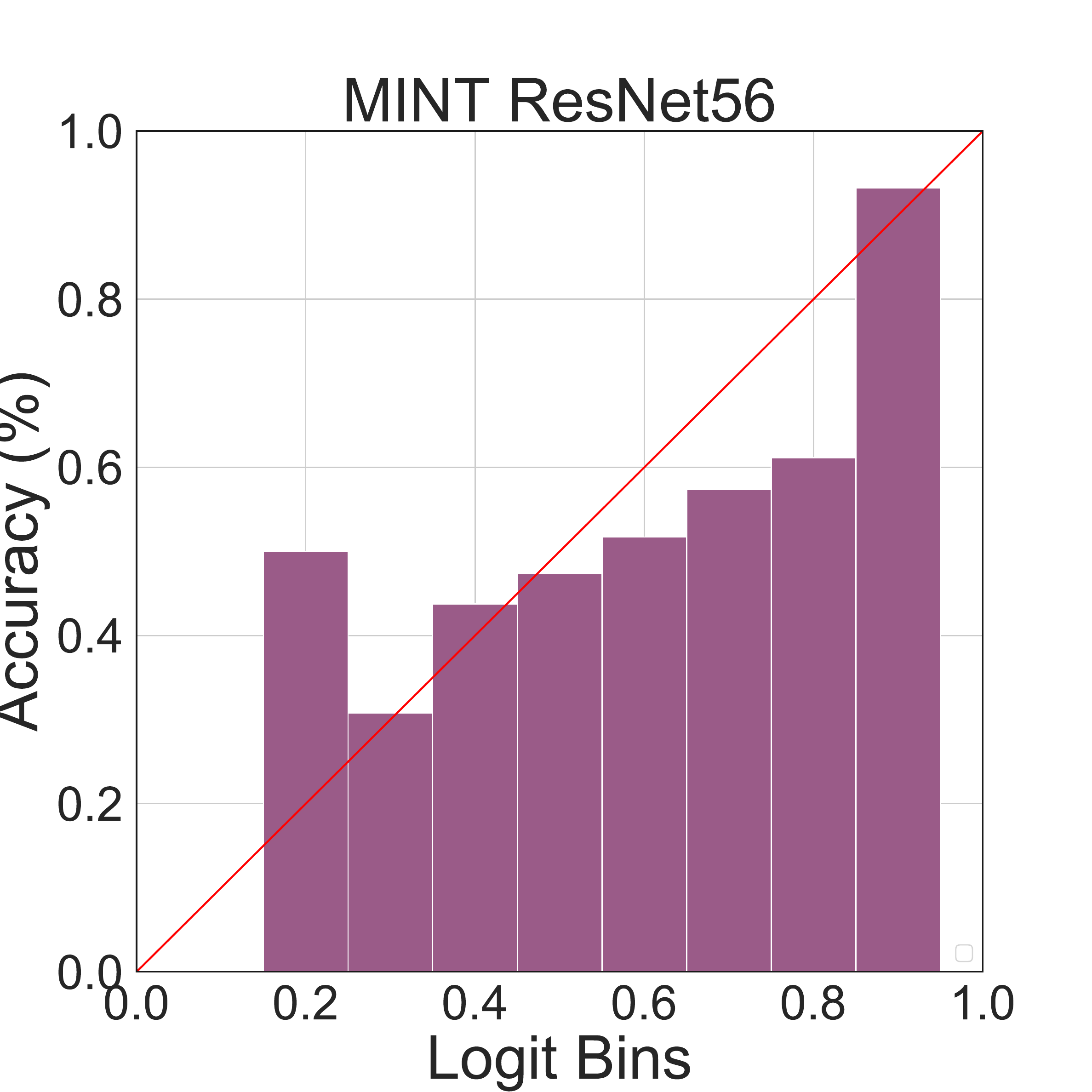}}
    \subfloat[ECE:0.0069]{\label{fig:cal_mint_resnet50}\includegraphics[width=0.25\textwidth]{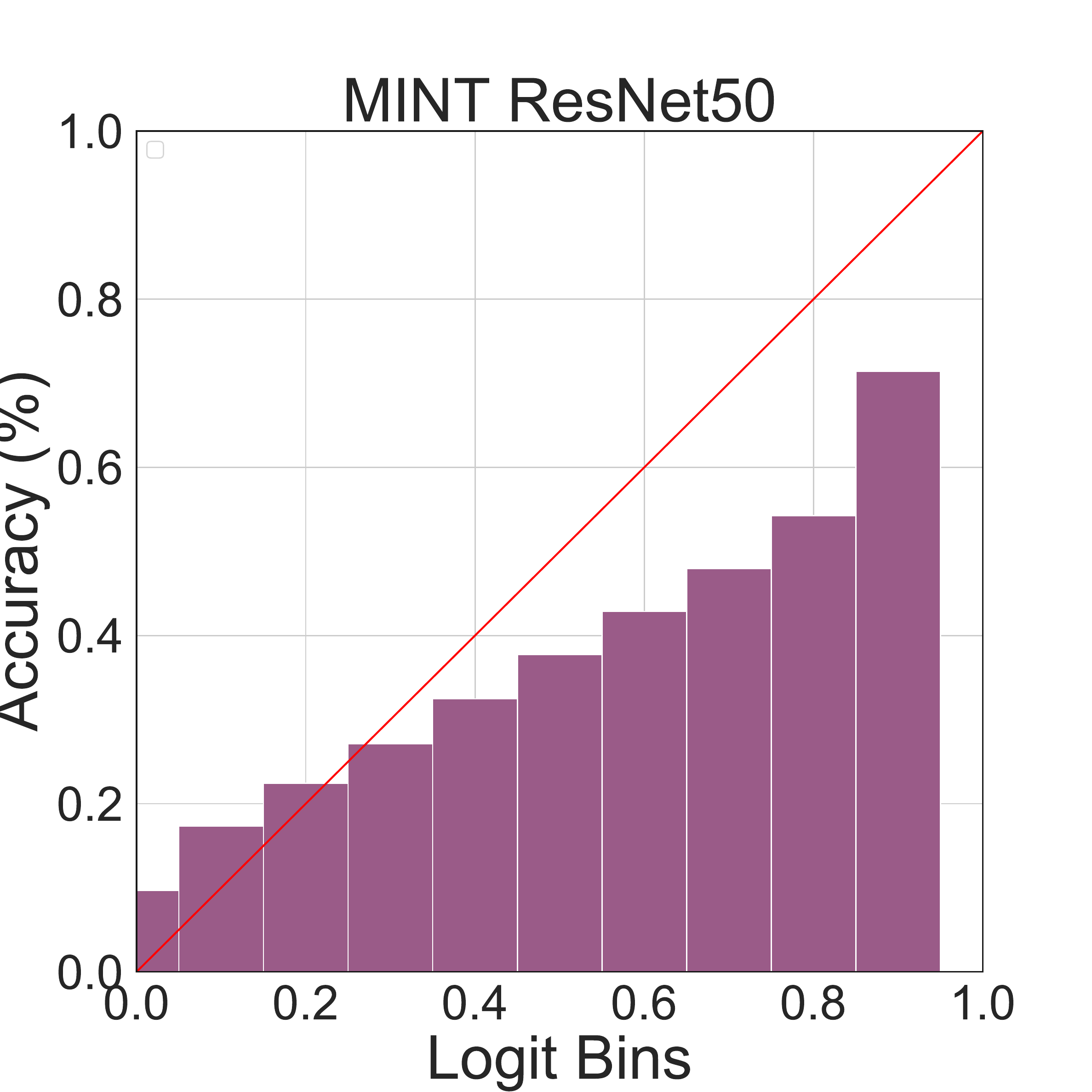}}
    \caption{Calibration statistics measure the agreement between the confidence output of the network and the true probability. The red line indicate the ideal trend if the confidence and probability matched. We observe that \alg-compressed networks act as a regularizer to decrease the Expected Calibration Error (ECE) when compared to the original network as well as better match the ideal curve}
    \label{fig:calibration}
\end{figure*}

\subsection{Comparison against existing methods}
\label{sec:comparison_against_existing_methods}

As a first step to showcasing \alg's abilities, we compare it against state-of-the-art baselines in network pruning.
The baselines in Table~\ref{tab:results_main} are arranged in ascending order of the percentage of parameters pruned.
Our algorithm clearly outperforms most of the SOTA pruning baselines across the number of pruned parameters while maintaining high accuracy and reducing the memory footprint of the network. 
We note that while some of the pruning baselines listed use multiple prune-retrain steps to achieve their result, we use only a \textbf{single step} to match and outperform them.

In Fig.~\ref{fig:sens_per_layer} we take a deeper look at how the overall compression $\%$ is spread throughout the network. 
Comparing Figs.~\ref{fig:sens_vgg16}, \ref{fig:sens_resnet56} and \ref{fig:sens_resnet50}, we can establish the strong influence of datasets and network architecture on where redundancies are stored. 
In the cases of VGG16 and ResNet56, training with CIFAR-10 leads to the storage of possibly redundant information in the latter portion of the networks.
The early portions of the network are extremely sensitive to pruning.
ResNet50 when trained on ILSVRC2012 forces compression to be more spread out across the network, possibly indicating the spread of redundant features at different levels of the network.

\subsection{Hyper-parameter Empirical Analysis}
\label{sec:hyper_parameter_empirical_analysis}
We take a closer look at two important hyper-parameters that help \alg~scale well to deep networks, (a) number of groups in a layer $G$, and (b) the number of samples per class, $m$, used to compute the conditional GMI.
Below, we look into how each of them impacts the maximum number of parameters pruned to achieve $> 98.50\%$ accuracy for MNIST on MLP.

\hfill \\
\noindent{\bf{Group size}}
$G$ directly corresponds to the number of filters that are grouped together when computing conditional GMI and thresholding.
More groups lead to lesser filters per group, which allows for more fine-grained computation of multivariate dependency and thereby, more precise pruning. 
In this experiment, $m = 250$.
Results in Fig.~\ref{fig:groups} match our expectation by illustrating the increase in the upper limit for parameters pruned to achieve the desired performance..

\hfill \\
\noindent{\bf{Samples per class}}
The number of samples per class directly impacts the final number of activations used to compute the conditional GMI.
The GMI estimator should improve its estimates as the number of samples per class and the total number of samples is increased.
In this experiment, $G = 20$.
Fig.~\ref{fig:samples} confirms our expectation by showing a steady improvement in the parameters pruned as the number of samples per class and thereby the total number of samples is increased.

\subsection{Characterization}
\label{sec:characterization}
The standard metric used to compare deep network compression methods is  the percentage of parameters pruned while maintaining recognition performance close to the baseline.
However, the original intent of compressing networks is to deploy them in real-world scenarios which necessitate  other characterizations like robustness to adversarial attacks and an ability to reflect true confidence in predictions.

To understand the impact of pruning networks in the context of adversarial attacks we use two common adversarial attacks, Iterative FGSM~\cite{goodfellow2014explaining}, which doesn't exclusively target a desired class, and Iterative-LL~\cite{kurakin2016adversarial}, which targets the selection of the least likely class.
Fig.~\ref{fig:adversarial_attacks} shows the response of other the original and \alg-compressed networks to both attacks.
We clearly observe that \alg-compressed networks are more vulnerable to targeted and non-targeted attacks.

While the core idea behind \alg~is to retain filters that contribute the majority of the information passed to the next layer, in using a subset of the available filters we remove a certain portion of the information passed down.
Fig.~\ref{fig:learned_representations} compares the portions of the image that contribute towards a desired target class, between the original (top row) and \alg-compressed networks (bottom row).
We observe that the use of a subset of filters in the compressed network has reduced the effective portions of the image that contribute towards a decision, not to mention minor modifications to the features used themselves.
We posit that the reduction in the number of filters used and the available redundant features is the reason for \alg-compressed networks being vulnerable to adversarial attacks.


Calibration statistics~\cite{naeini2015obtaining,10.5555/3305381.3305518} measure the agreement between the confidence provided by the network and the actual probability.
These measures provide an orthogonal perspective to adversarial attacks since they measure statistics only for in-domain images while adversarial attacks alter the input.
Fig.~\ref{fig:calibration} highlights the decrease in Expected Calibration Error (ECE) for the \alg-compressed networks when compared to their original counterparts.
The plot illustrates that the histogram trend is closer to matching the ideal trend indicated by the linear red curve.
After pruning, the sparse networks seem to behave similarly to a regularizer by focusing on a smaller subset of features and decreasing the ECE.
On the other hand, the original networks contain many levels of redundancies which translates to overfitting and having higher ECE.


\section{Conclusion}
\label{sec:conclusion}
In this work, we propose \alg~as a novel approach to network pruning in which the dependency between filters of successive layers is used as a measure of importance. 
We use conditional GMI to evaluate importance and incorporate stochasticity in our algorithm in order to retain filters that pass the majority of the information through layers.
In doing so, \alg~achieves better pruning performance than SOTA baselines, using a single prune-retrain step.
When characterizing the behaviour of \alg-pruned networks, we observe that it behaves like a regularizer and improves the calibration of the network.
However, a reduction in the number of filters used and redundancies makes pruned networks susceptible to adversarial attacks.
Our future direction of work includes improving the robustness of compressed networks to adversarial attacks as well as detailing the sensitivity of layers in various architectures to network pruning.

\section*{Acknowledgment}
This work has been partially supported (Madan Ravi Ganesh and Jason J. Corso) by NSF IIS 1522904 and NIST 60NANB17D191 and (Salimeh Yasaei Sekeh) by NSF 1920908; the findings are those of the authors only and do not represent any position of these funding bodies.
The authors would also like to thank Stephan Lemmer for his invaluable input on the calibration of deep networks.

\newpage
%
%
\bibliographystyle{IEEEtran}
\bibliography{egbib}


\clearpage
\appendix
\subsection{Practical Implementation Hyper-parameter}
\label{sec:practical_implementation}

\begin{figure}[b!]
    \renewcommand\thefigure{A-1}
    \centering
    \includegraphics[width=\columnwidth]{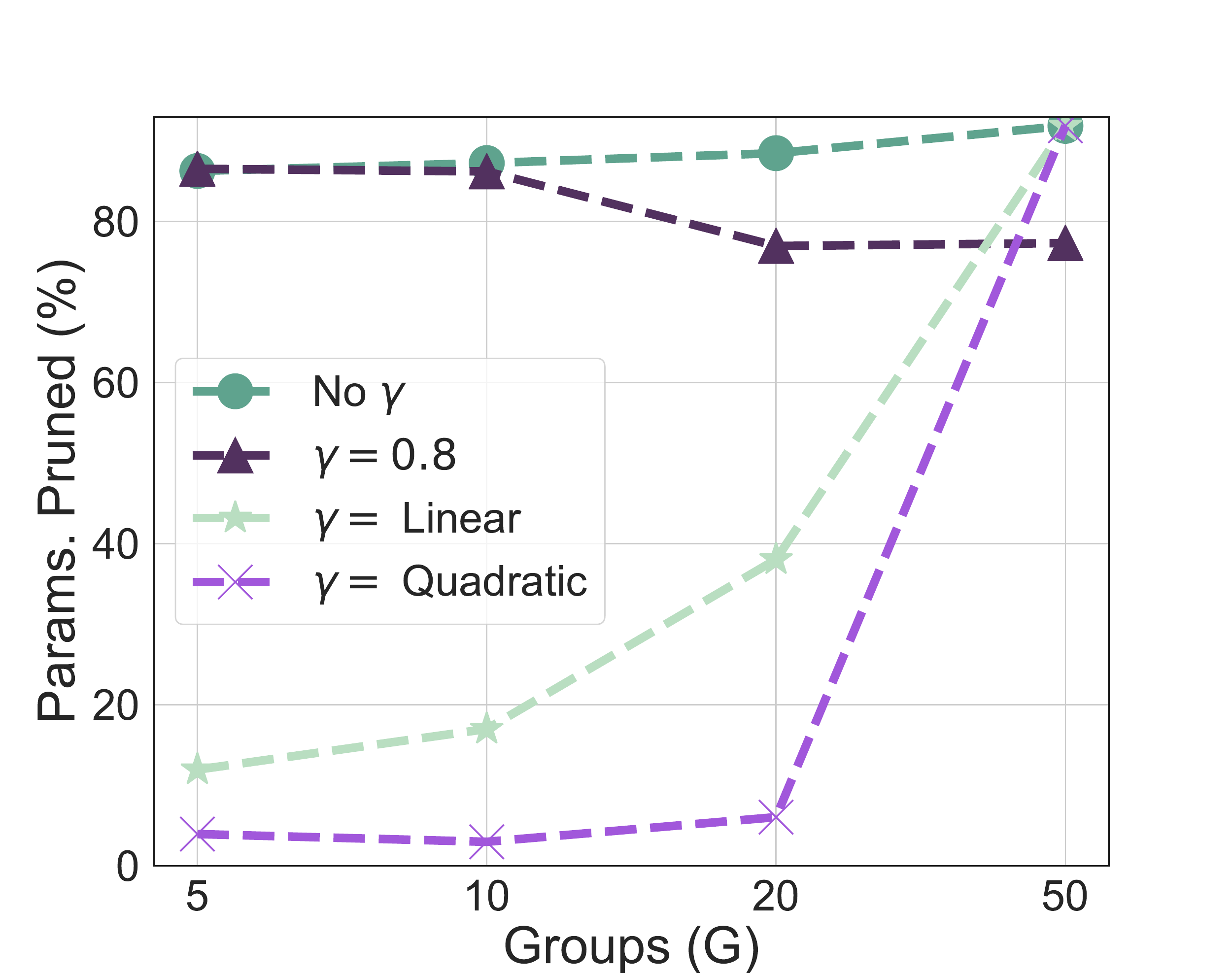}
    \caption{Varying $\gamma$ using a linear or quadratic dependence on $G$ shows a positive correlation to parameters pruned while using a constant value forces irregular behaviour}
    \label{fig:gamma}
\end{figure}
In Section 3.2 of the main paper, we describe \alg~in detail. 
In the description, we use $\delta$ as a threshold on $\rho()$ values to help retain only those dependencies that contribute a majority of the information to the next layer.
However, in practice, since the $\rho()$ values are not normalized w.r.t. each other, there are cases when certain layers can be fully removed when the $\delta$ parameter is large. 
In order to protect the network from such behaviour, we use an additional threshold called $\gamma$ to cap the maximum percentage of filters that can be removed from a selected layer.
If the pruning ratio within a layer exceeds the value of $\gamma$, then $\gamma$ is used to re-evaluate $\delta$ such that it falls within the desired pruning ratio.
The new temporary value of $\delta$ is computed using the $\rho()$ values within the pair of layers considered, and is only used prune filters in the selected pair of layers.

Fig.~\ref{fig:gamma} illustrates the impact of $\gamma$ on the percentage of parameters pruned in the MNIST-MLP experimental setup.
Here, we observe a relatively steady increase in the maximum number of parameters pruned as the number of groups ($G$) increases, when no $\gamma$ is provided or when a linear/quadratic dependence on $G$ is used to compute $\gamma$.
However, when $\gamma$ is fixed at 0.8, we observe irregular behaviour, where the maximum percentage of parameters removed decreases with an increase in $G$.
We posit that this behaviour arises from two factors, (a) the varied sensitivity of different layers is not compensated for by a fixed $\gamma$ value, and (b) the difference in $G$ values leads to different GMI estimates which account for more uncertainty in measurement.

\subsection{Experimental Setup}
\label{sec:experimental_setup}
In the following section we outline the hyper-parameters used during each stage of our experimental pipeline, training, pruning and retraining.
The values are provided here in an effort to help users replicate our experiments. 
We plan to release our code soon.

\noindent{\bf{Training Setup:}}
Table~\ref{tab:training} outlines the setups used to obtain trained models from which mutual information (MI) estimates are computed. 
A standard pretrained model from PyTorch zoo is used for the ResNet50-ImageNet experiments.

\begin{table}[t]
\renewcommand\thetable{B-1}
\centering
\caption{Training setups used to obtain pre-trained network weights}
\begin{tabular}{lccc}
\toprule
       & MLP & VGG16  & ResNet56  \\ \midrule
Epochs                            & 30          & 300             & 300                                       \\
Batch Size                        & 256         & 128             & 128                                       \\
Learning Rate                     & 0.1         & 0.1             & 0.01                                      \\
Schedule                          & 10, 20      & 90, 180, 260    & 150, 225                                  \\
Optimizer                         & SGD         & SGD             & SGD                                       \\
Weight Decay                      & 0.0001      & 0.0005          & 0.0002                                    \\
Multiplier                        & 0.1         & 0.2             & 0.1                                        \\
\bottomrule
\end{tabular}
\label{tab:training}
\end{table}

\begin{table}[t]
\renewcommand\thetable{B-2}
\centering
\caption{Retraining setups used to obtain final performance listed in Table 1 of main paper}
\begin{tabular}{lcc}
\toprule
       & MLP  & VGG16   \\ \midrule
Epochs         & 30          & 300                          \\
Batch Size     & 256         & 128                      \\
Learning Rate  & 0.1         & 0.1                       \\
Schedule       & [10, 20]    & [90, 180, 260]                                  \\
Optimizer      & SGD         & SGD                           \\
Weight Decay   & 0.0001      & 0.0005                           \\
Multiplier     & 0.1         & 0.2                            \\
\bottomrule
\end{tabular}
\label{tab:retraining}
\end{table}

\begin{table}[t!]
\renewcommand\thetable{B-3}
\centering
\caption{Retraining setups used to obtain final performance listed in Table 1 of main paper contd}
\begin{tabular}{lcc}
\toprule
       &  ResNet56 & ResNet50 \\ \midrule
Epochs         & 300  & 130                                      \\
Batch Size     & 128  & 64                                   \\
Learning Rate  & 0.1 & 0.1                                    \\
Schedule       & [90, 180, 260]  & [30, 60, 90, 100]                                \\
Optimizer      & SGD      & SGD                                   \\
Weight Decay   & 0.0002   & 0.0001                                   \\
Multiplier     & 0.1      & 0.1                                    \\
\bottomrule
\end{tabular}
\label{tab:retraining_contd}
\end{table}

\noindent{\bf{Pruning Setup:}}
For the conditional geometric mutual information (GMI) estimate we provide a set of parameters describing the number of groups per layer, $G$ as well as the number of samples per class, $m$.
We use the average activation across a filter's dimensions as the samples for GMI computations.

\noindent{\bf{MLP - MNIST:}}
Between FC 1 and FC2, $G$ is set to 250 and 100 while between FC 2 and FC 3 it is set to 300 and 10. 650 samples per class are used to compute the conditional GMI estimates.

\noindent{\bf{VGG16 - CIFAR10:}}
All the layers between Convolution 4 and Linear 1 use $G = 64$ while $m = 650$.
$\gamma=0.45$ is used for Convolution layers 5, 6, 7, and 8.

\noindent{\bf{ResNet56 - CIFAR10:}}
$m$ is set to 500 samples per class.
All the layers between Convolution 1 and Convolution 19 use $G = 16$, Convolution 20 to 37 use $G = 32$ while layers up to Convolution 55 use $G = 64$.
We skip pruning Convolution layer 16, 20, 38, and 54.

\noindent{\bf{ResNet50 - ILSVRC2012:}}
$m$ is set to 5 samples per class.
All the convolution layer use $G = 64$.
We skip pruning the final linear layer.
\hfill \\

\noindent{\bf{Retraining Setup:}}
Tables~\ref{tab:retraining} and \ref{tab:retraining_contd} outline the setups used to obtain the final values used in Table 1 of the main paper. 
While retraining the MLP took approximately 1-2 minutes, VGG16 took about 1.5 hours, ResNet56 took 2.5 hours and ResNet50 on ILSVRC2012 took close to a week on 1 GPU.



%


\end{document}